\documentclass[11pt]{article}

\usepackage[final]{acl}
\usepackage{cuted}
\usepackage{times}
\usepackage{latexsym}
\usepackage{booktabs}
\usepackage[T1]{fontenc}
\usepackage{makecell}

\usepackage{tabularx}

\usepackage[utf8]{inputenc}

\usepackage{microtype}

\usepackage{inconsolata}
\usepackage{cuted}
\usepackage{graphicx}

\usepackage{amsmath}
\usepackage{amssymb}
\usepackage{array}
\usepackage{xcolor}      
\usepackage{colortbl}    
\usepackage{multirow}    
\usepackage{booktabs}
\usepackage{pifont}
\usepackage{cleveref}
\usepackage{tcolorbox}
\usepackage{makecell}
\usepackage{svg}
\usepackage{siunitx}
\usepackage{arydshln}
\usepackage{threeparttable}
\usepackage{wrapfig}
\usepackage[table]{xcolor}
\usepackage{hyperref}
\usepackage{url}
\usepackage{float}
\usepackage{placeins}
\usepackage{fontawesome5}
\usepackage{enumitem}
\usepackage[ruled,vlined,linesnumbered]{algorithm2e}
\usepackage{tikz}
\usepackage{stfloats}
\newcolumntype{L}[1]{>{\raggedright\arraybackslash}m{#1}}
\newcommand{\openailogo}{\includegraphics[height=1.2em]{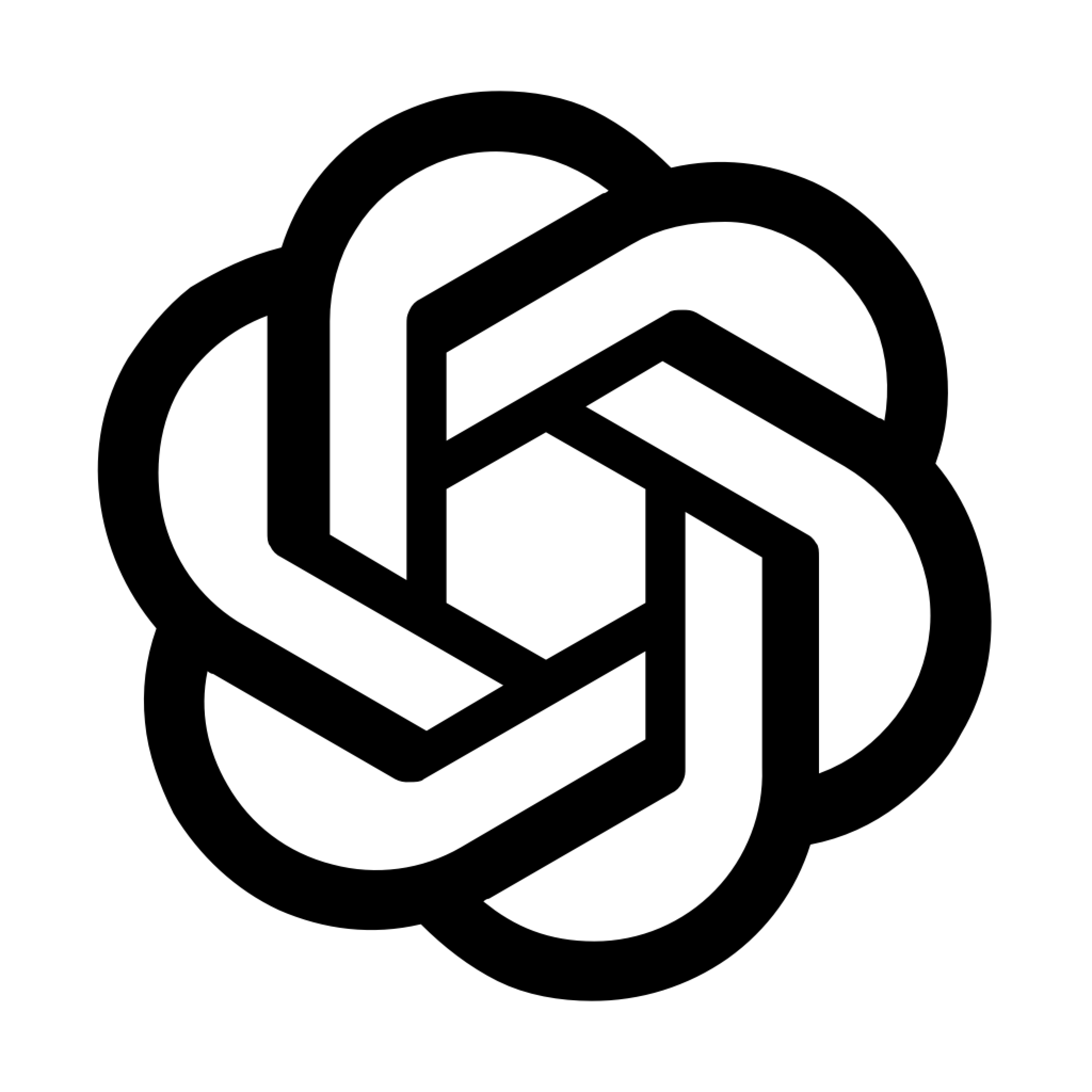}}
\newcommand{\qwenlogo}{\includegraphics[height=1.2em]{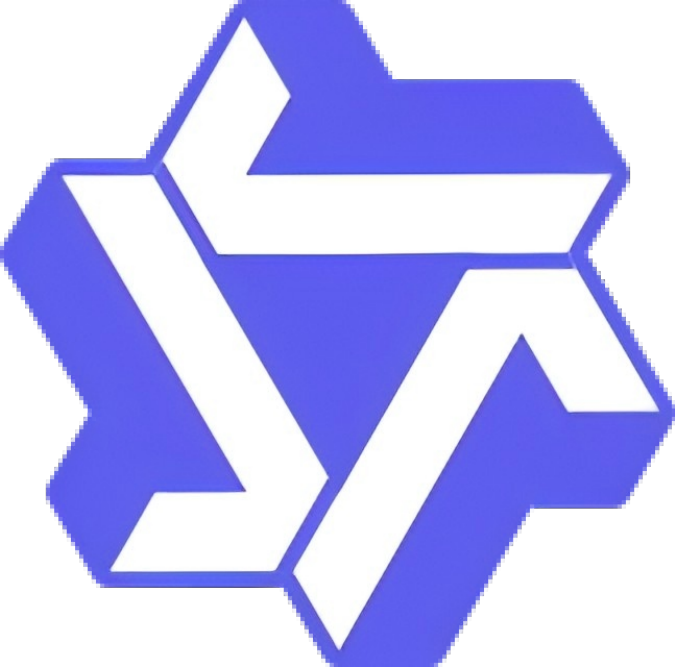}}
\newcommand{\memorybench}[1]{\textsc{MemoryBench}}
\newcommand{\longmemeval}[1]{\textsc{LongMemEval}}
\newcommand{\ours}[1]{TSM}

\definecolor{gptgray}{RGB}{245,246,248}
\definecolor{gptbar}{RGB}{230,232,236}
\definecolor{qwenlav}{RGB}{244,240,252}
\definecolor{qwenbar}{RGB}{228,220,246}

\title{Beyond Dialogue Time:\\ Temporal Semantic Memory for Personalized LLM Agents}

\author{
 \textbf{Miao Su\textsuperscript{1,2,3,4}},
 \textbf{Yucan Guo\textsuperscript{1,2,3}},
 \textbf{Zhongni Hou\textsuperscript{4}},
 \textbf{Long Bai\textsuperscript{1,2}},
\\
 \textbf{Zixuan Li\textsuperscript{1,2}},
 \textbf{Yufei Zhang\textsuperscript{4}},
 \textbf{Guojun Yin\textsuperscript{4}},
 \textbf{Wei Lin\textsuperscript{4}},
\\
 \textbf{Xiaolong Jin\textsuperscript{1,2,3}\thanks{\ \ Corresponding author.}},
 \textbf{Jiafeng Guo\textsuperscript{1,2,3}},
 \textbf{Xueqi Cheng\textsuperscript{1,2,3}},
\\
 \textsuperscript{1}Institute of Computing Technology, Chinese Academy of Sciences\\
 \textsuperscript{2}State Key Laboratory of AI Safety\\
 \textsuperscript{3}School of Computer Science, University of Chinese Academy of Sciences\\
 \textsuperscript{4}Meituan
\\
 \small{
    \href{mailto:sumiao22z@ict.ac.cn}{sumiao22z@ict.ac.cn}
 }
}

\begin{document}
\maketitle
\begin{abstract}
Memory enables Large Language Model (LLM) agents to perceive, store, and use information from past dialogues, which is essential for personalization.
However, existing methods fail to properly model the temporal dimension of memory in two aspects: 1) Temporal inaccuracy: memories are organized by dialogue time rather than their actual occurrence time; 2) Temporal fragmentation: existing methods focus on point-wise memory, losing durative information that captures persistent states and evolving patterns. 
To address these limitations, we propose Temporal Semantic Memory (TSM), a memory framework that models semantic time for point-wise memory and supports the construction and utilization of durative memory.
During memory construction, it first builds a semantic timeline rather than a dialogue one. Then, it consolidates temporally continuous and semantically related information into a durative memory. During memory utilization, it incorporates the query’s temporal intent on the semantic timeline, enabling the retrieval of temporally appropriate durative memories and providing time-valid, duration-consistent context to support response generation.
Experiments on \textsc{LongMemEval} and \textsc{LoCoMo} show that \ours{} consistently outperforms existing methods and achieves up to 12.2\% absolute improvement in accuracy, demonstrating the effectiveness of the proposed method.
Code is available at \url{https://github.com/cosmicexotic/TSM}.

\end{abstract}

\section{Introduction}

\begin{figure}[t]
\centering

    \includegraphics[width=\linewidth]{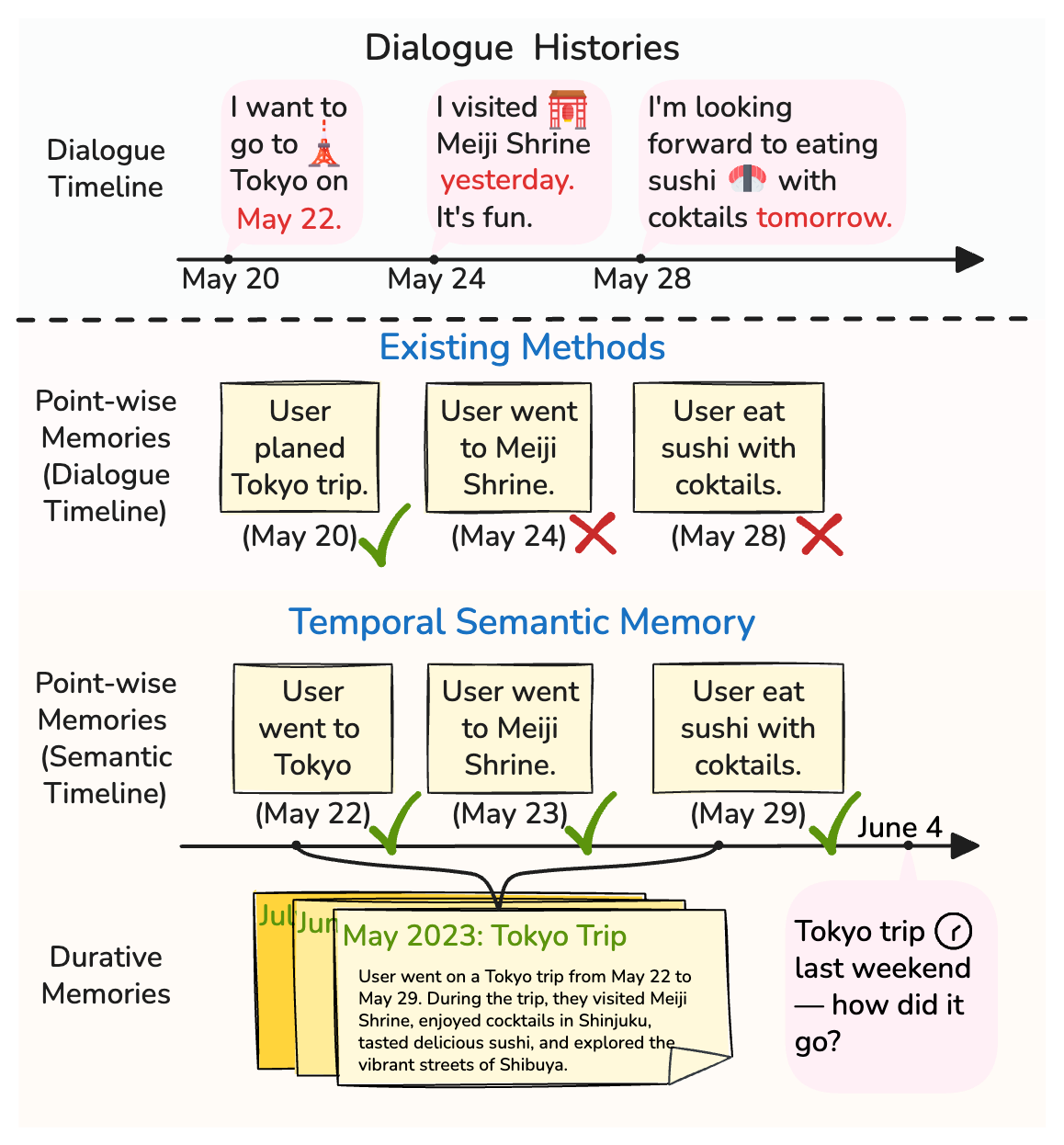}
    \caption{Comparison of existing methods vs. \ours{} with semantic timeline and durative memories.}
    \label{fig:motivation}
\end{figure}

Recent years have witnessed the rapid emergence of Large Language Model (LLM) agents, autonomous systems built upon LLMs with capabilities for reasoning, tool use, and long-term interaction~\cite{matarazzo_survey_2025,minaee_large_2025,luo_large_2025}. 
A key component of LLM agents is \textit{memory}. Instead of changing model parameters, memory provides an explicit store of interaction-derived information that agents can later reuse, enabling adaptation over extended interactions~\cite{sumers2023cognitive,wu_human_2025}. 
Personalized dialogue agents~\cite{chhikara2025mem0, li_hello_2025} exemplify this setting. Memory helps the agent maintain user-specific context, such as past plans, preferences, and evolving situations. 
It is typically implemented as a three-stage pipeline of construction, update, and utilization~\cite{hu_memory_2025}, so that responses remain coherent, grounded in prior interactions, and consistent over time.

Recent works construct memory by organizing dialogue histories into structured memory entries and retrieving them when needed~\cite{zhong_memorybank_2023, kim_pre-storage_2025, tan_prospect_2025, sun_hierarchical_2025}. 
However, as shown in~\Cref{fig:motivation}, existing methods suffer from two critical limitations in how they model temporal information.
(1) \textbf{Temporal inaccuracy.} Most systems treat the \emph{dialogue timeline} (when a chat turn is produced) as the primary temporal signal when assessing recency or relevance~\cite{rasmussen_zep_2025}. 
This is problematic because users often talk about events that occur at different times than the conversation itself, including future plans, past trips, and ongoing states. 
For example, when a user talks on May 28 about a trip happening on May 29, dialogue time and event time are misaligned; using the dialogue timeline alone can cause the system to store or retrieve memories under the wrong time context.
(2) \textbf{Temporal fragmentation.} Many methods store memories as isolated, point-wise entries~\cite{tan_prospect_2025, fang_lightmem_2025,chhikara_mem0_2025,rasmussen_zep_2025}. 
This representation breaks temporally continuous experiences into disconnected records, making it difficult to recover durations and long-term states.
For instance, the multiple entries within a week in~\Cref{fig:motivation} together describe a coherent Tokyo trip; treating them independently ignores their temporal continuity and semantic relatedness, which in turn hinders the formation of persistent states and evolving patterns.
Together, these issues prevent agents from retrieving complete and relevant memories, especially when the user query implicitly assumes a coherent real-world timeline.

In contrast, human memory relies on time as a scaffold for ordering and linking real-life experiences, supporting coherent recall across long-term memories~\cite{macdonald_hippocampal_2011,huet_episodic_2025}.
This highlights the importance of modeling time beyond a point-wise, dialogue-timeline view.

To this end, we propose \textbf{Temporal Semantic Memory (TSM)}, a memory framework designed to support semantic-time grounded and duration-aware memory access.
In particular, \ours{} addresses the above challenges with two tightly coupled components:
(1) \textbf{Duration-aware memory construction.} TSM builds a \emph{semantic timeline} through a temporal knowledge graph, aligning memory with when events happen and how long they last.
Beyond recording point-wise facts, \ours{} links temporally continuous and semantically related mentions and consolidates them into \emph{durative summaries} that capture long-term states (i.e., topics and personas).
(2) \textbf{Semantic-time guided memory utilization.} During memory utilization, \ours{} incorporates the query's semantic temporal intent and retrieves memories at the appropriate temporal granularity, rather than relying on dialogue-time recency or semantics-only similarity that disregards timing.
This enables the system to return time-valid, duration-consistent context for response generation with correct temporal grounding.
In addition, \ours{} maintains memory with a lightweight hierarchical mechanism: it incrementally updates temporal facts online and periodically consolidates summaries to improve long-term consistency.

Extensive experiments on \textsc{LongMemEval}~\cite{wu_longmemeval_2025} and \textsc{LoCoMo}~\cite{maharana_evaluating_2024} demonstrate that \ours{} consistently outperforms strong memory baselines, with the largest gains on multi-session understanding and temporal reasoning tasks, validating the effectiveness of semantic-time grounding and duration-aware consolidation.

\section{Related Work} 

\subsection{Agent Memory}

LLM agents are increasingly equipped with long-term memory that grows and adapts over time, allowing them to accumulate knowledge, recall prior context, and adjust behavior based on experience~\citep{owl2025, openmanus2025, google2025deepresearch, bytedance2025deerflow}. There are several functions of agent memory: 
Factual memory stores persistent information such as user profiles, dialogue history, and world facts to support long-term consistency and personalization~\cite{wu2026knowme}, as explored in memory-augmented dialogue agents and long-term user modeling systems~\citep{park_generative_2023, packer_memgpt_2024,nan_nemori_2025,kwon_embodied_2025};
experiential memory records past interaction trajectories and distilled strategies to enable continual self-improvement across tasks, exemplified by case-based, strategy-based, and skill-based learning in reflective and self-improving agents~\citep{zhang2026expseek, shinn2023reflexion,yan_memory-r1_2025,zhou_mem1_2025,ouyang_reasoningbank_2025};
working memory provides mechanisms for the active management of transient context~\cite{zhou_mem1_2025,zhang_memgen_2025}.

In this work, we focus on user-specific factual memory to support temporally grounded, personalized agent over long-term interactions.

\subsection{Graph-structured Memory}
In the context of agent memory, graph-structured memory arises naturally when agents accumulate relational insights over time~\cite{yang2026graph}. 
Mem0$^{g}$ employs a scalable two-phase architecture (extraction and update) that dynamically stores and retrieves salient facts in a knowledge graph~\cite{chhikara2025mem0}.
A-MEM~\cite{xu2025mem} builds an interconnected memory network inspired by Zettelkasten~\cite{kadavy_digital_2021}, where each new memory is represented as a structured note with attributes (e.g., keywords/tags), and the system dynamically links related memories and updates existing notes as new information arrives, enabling an evolving memory graph.
Zep~\cite{rasmussen_zep_2025} proposes a temporal knowledge graph memory layer, emphasizing temporal reasoning over an evolving graph rather than static document retrieval. However, it ignores time during the memory retrieval stage. Other recent systems~\cite{zhang_g-memory_2025, wuGAMHierarchicalGraphbased2026a} similarly allow the graph to be constructed, extended, or reorganized throughout the agent's operation.

Many prior graph-based memory systems primarily treat the graph as a persistent store of extracted facts, but the \emph{evolution} of memory in a time period is not considered, yet the retrieval stage still underutilizes the \emph{semantic time} encoded in the graph, resulting in temporally misaligned recall.

\section{Methodology}

\begin{figure*}
    \centering
    \includegraphics[width=\textwidth]{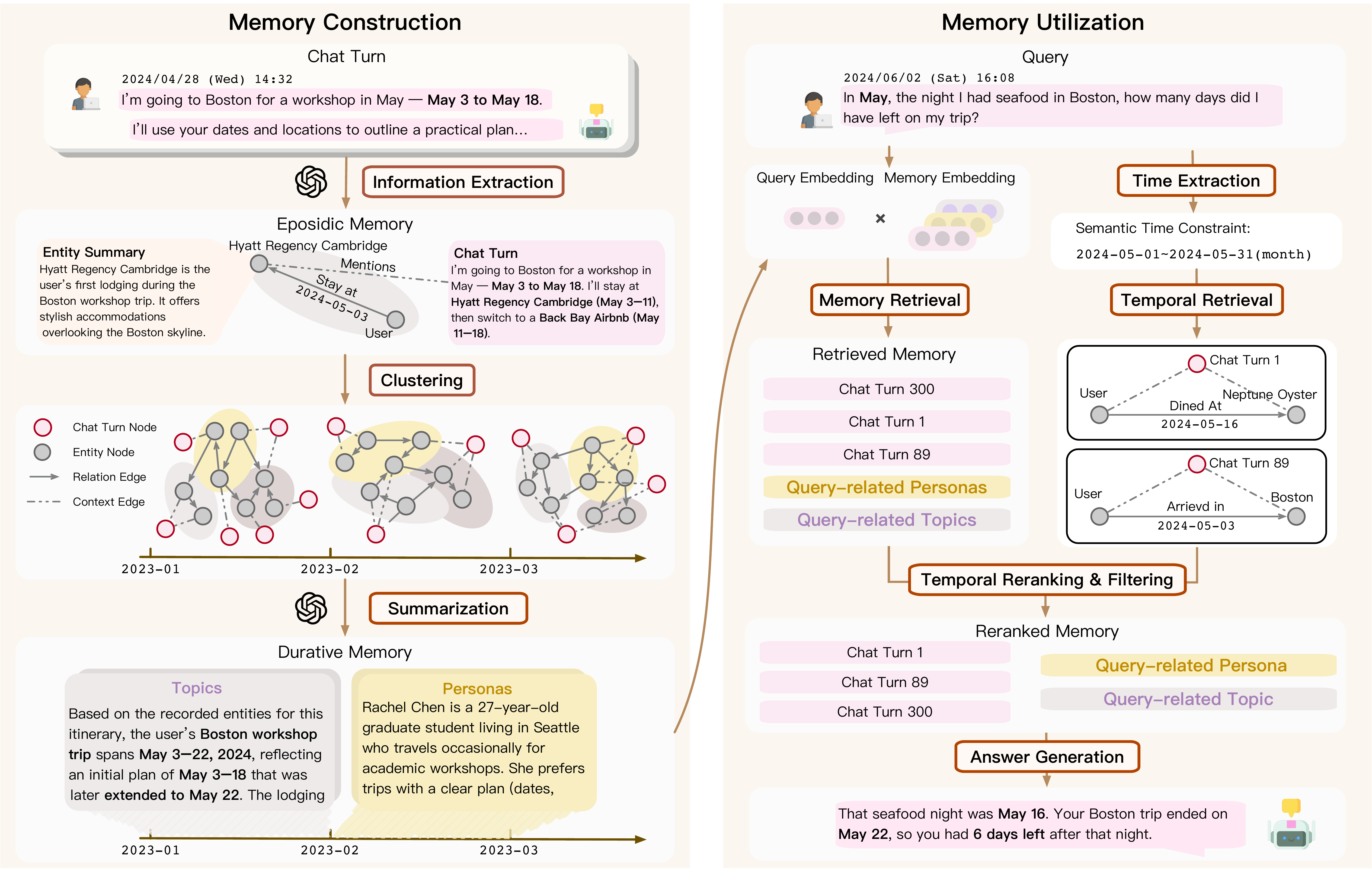} 
    \caption{The overall framework of \ours{}. Memory consolidation constructs a temporal knowledge graph from episodic memory and subsequently consolidates it into time-aware durative memory. Memory utilization retrieves accurate memories by applying semantic-temporal constraints.} 
    \label{fig:framework}
    
\end{figure*}

\subsection{Preliminary}
We consider the task of building a personalized dialogue agent in a \textbf{multi-session} conversational setting.
A \textit{session} represents a distinct interaction period, often delimited by user inactivity, explicit user confirmation of conversation completion, or the initiation of a new dialogue thread.
Within each session, the conversation unfolds as a sequence of chat turns, where a \textit{chat turn} consists of a user query and the agent's corresponding response.


Typically, a standard agent memory system implements three core functions: 

(1) Construction, which transforms raw chat history $D$ into structured memory $M$,
\[
M = f_{\text{construct}}(D)
\]

(2) Update, which refines the existing memory with new conversational data $D'$,

\[
M' = f_{\text{update}}(M,D')
\]

(3) Utilization (Retrieval), which generates an answer $A$ to a given query $Q$ based on the stored memory,
\[
A = f_{\text{retrieve}}(M,Q).
\]

\subsection{Overview}


As illustrated in Figure~\ref{fig:framework}, \ours{} consists of two stages: Memory Construction (\S\ref{sec:construction}) that builds a temporal knowledge graph as a episodic memory and organizes episodic interactions into durative memories; Memory Utilization (\S\ref{sec:retrieval}) that performs constraint-aware retrieval by integrating dense matching with temporal reranking and filtering for accurate memory access. Additionally, we build an Update Mechanism (\S\ref{sec:update}) that periodically refreshes both the episodic memory and the constructed durative memory to ensure long-term consistency. 




\subsection{Duration-aware Memory Construction}
\label{sec:construction}
We construct two complementary memory types from dialogue streams through a hierarchical process:

\noindent\textbf{Episodic Memory} captures atomic facts with specific semantic timestamp (e.g., "visited Tokyo on May 23"), organized as a temporal knowledge graph with explicit temporal grounding.

\noindent\textbf{Durative Memory} maintains enduring patterns that persist across extended periods, derived from episodic memory through temporal segmentation and semantic abstraction. Unlike event-specific episodic memories, durative memories represent stable user characteristics (e.g., sustained interests, evolving preferences) extracted from accumulated experiences. This concept parallels semantic memory in cognitive science.

\subsubsection{Episodic Memory Construction}

We construct a Temporal Knowledge Graph (TKG) as a structured memory index that records episodic information mentioned in the dialogue history following Zep~\cite{rasmussen_zep_2025-1}. The TKG is not directly used as retrievable memory content; instead, it provides precise temporal localization and semantic time access for subsequent memory construction and retrieval.

Formally, the TKG is defined as a set of temporally grounded facts
\begin{equation}
\mathcal{G} = \{ (e_s, r, e_o, t) \mid t \in \mathbb{T} \}.
\end{equation}
where $e_s$ and $e_o$ denote the subject and object entities, $r$ denotes a semantic relation, and $t$ denotes the time point when the fact is valid.

In addition to temporal facts, each entity node maintains a compact entity summary extracted from its supporting dialogue contexts. We represent an entity as $e \triangleq (n_e, s_e),$ where $n_e$ is the canonical entity name and $s_e$ is an LLM-generated entity summary that consolidates salient attributes of $e$.

We extract entities and relations from each turn (with a context window of the preceding $n$ turns) and incrementally integrate them into the TKG via deduplication and temporal consistency checks (details in Section~\ref{sec:update}).

The resulting TKG serves as an episodic memory index that supports time-aware access, temporal filtering, and consistency checking, while deferring semantic abstraction to later stages.

\subsubsection{Durative Memory Construction}

Based on the episodic memory graph, we construct durative memory by aggregating episodic information into higher-level semantic representations. 

Given the temporal knowledge graph, we partition it into a sequence of temporal slices
\begin{equation}
\begin{aligned}
\mathcal{G} &= \bigcup_{k} \mathcal{G}^{(k)}, \\
\mathcal{G}^{(k)} &= \{ (e_s, r, e_o, t) \mid t \in [\tau_k, \tau_{k+1}) \},
\end{aligned}
\end{equation}
where $[\tau_k, \tau_{k+1})$ denotes a fixed temporal interval. In our implementation, the granularity is set to one month by default.

For each temporal slice $\mathcal{G}^{(k)}$, we collect the involved entity set
\begin{equation}
\mathcal{E}^{(k)} = \{ e \mid e \text{ appears in } \mathcal{G}^{(k)} \}.
\end{equation}

We apply a Gaussian Mixture Model (GMM)~\cite{huang_retrieval-augmented_2025} to cluster entities within the same temporal slice
\begin{equation}
p(z \mid e) = \mathrm{GMM}(\mathbf{h}^{\mathrm{name}}_e),
\end{equation}
where $z$ denotes a latent cluster capturing a coherent semantic theme. For each entity $e$, we assign it to the most likely cluster $a(e)=\arg\max_{z} p(z\mid e)$, and define
\begin{equation}
\mathcal{E}_z = \{ e \in \mathcal{E}^{(k)} \mid a(e)=z \}.
\end{equation}

For each cluster $z$ in the $k$-th temporal slice, let
\begin{equation}
\mathcal{X}_z = \{ (n_e, s_e) \mid e \in \mathcal{E}_z \}
\end{equation}
denote the entity-summaries in the cluster. We define a topic as
\begin{equation}
\begin{aligned}
\mathrm{Topic}_z &= \{ \tau_k, s_z, \mathbf{c}_z \}, \\
s_z &= \mathrm{LLM}_{\mathrm{sum}}(\mathcal{X}_z), \\
\mathbf{c}_z &= \mathrm{Embedding}(s_z),
\end{aligned}
\end{equation}
where $\tau_k$ denotes the corresponding time slice, $s_z$ is the textual topic summary, and $\mathbf{c}_z$ is its embedding used for downstream retrieval.

To capture user-level characteristics, we further aggregate the dialogue contexts associated with the entities in $\mathcal{E}_z$. Using a bidirectional index between entities and their originating chat turns, we collect the corresponding dialogue
\begin{equation}
\mathcal{D}_z = \{ d \mid d \text{ mentions } e,\; e \in \mathcal{E}_z \}.
\end{equation}

The persona representation is defined as
\begin{equation}
\begin{aligned}
\mathrm{Persona}_z &= \{ \tau_k, \mathbf{p}_z, \mathbf{u}_z \}, \\
\mathbf{p}_z &= \mathrm{LLM}_{\mathrm{sum}}(\mathcal{D}_z), \\
\mathbf{u}_z &= \mathrm{Embedding}(\mathbf{p}_z),
\end{aligned}
\end{equation}
where $\mathbf{p}_z$ captures stable user traits, preferences, and behavioral patterns expressed within the temporal slice, and $\mathbf{u}_z$ denotes its embedding.

Through temporal segmentation and semantic abstraction, the constructed topics and personas form hierarchical, temporally anchored durative memory. It captures durative user states beyond isolated point events, thereby supporting efficient long-term storage and subsequent constraint-aware retrieval.

\subsection{Semantic-time Guided Memory Utilization}
\label{sec:retrieval}
This stage retrieves memory that is both semantically relevant and temporally consistent with the user query. Given a query $q$, we (i) infer its semantic time constraint $T_q$, (ii) perform dense retrieval over topics, personas, and raw dialogue chunks, and (iii) enforce the temporal constraint by filtering time-anchored summaries and promoting candidates supported by temporally valid evidence from the TKG.

Given a user query $q$ issued at time $t_{\mathrm{now}}$, we first parse its semantic-time constraint $T_q$, i.e., the time range when the described event is intended to hold (rather than the dialogue time)
\begin{equation}
T_q = \mathrm{ParseTime}(q, t_{\mathrm{now}}),
\end{equation}
where $\mathrm{ParseTime}(\cdot)$ is implemented with spaCy~\cite{Honnibal2020spaCy} and resolves both explicit and relative time expressions.

Let the retrievable memory pool be
\begin{equation}
\mathcal{M} = \mathcal{M}_{\mathrm{topic}} \cup \mathcal{M}_{\mathrm{persona}} \cup \mathcal{M}_{\mathrm{raw}},
\end{equation}
where each topic/persona entry $m$ carries a slice timestamp $\tau(m)$ from construction, while each raw segment $m \in \mathcal{M}_{\mathrm{raw}}$ is a chat turn. We compute dense retrieval scores 
\begin{equation} s_{\mathrm{sem}}(m;q) = \mathrm{sim}\big(\mathrm{Enc}(q), \mathrm{Enc}(m)\big), \end{equation} 
and retrieve Top-$K$ candidates \begin{equation} \mathcal{D} = \mathrm{TopK}_{m \in \mathcal{M}} \; s_{\mathrm{sem}}(m;q). 
\end{equation}

To align retrieval with the temporal intent in $q$, we apply temporal filtering to the retrieved topics/personas (post-retrieval in our implementation):
\begin{equation}
\mathrm{Keep}(m,T_q)=
\begin{cases}
\mathbb{I}\!\big[\tau(m) \in T_q\big], &
\begin{aligned}[t]
m \in \mathcal{M}_{\mathrm{topic}}\\
\cup\ \mathcal{M}_{\mathrm{persona}},
\end{aligned}\\[4pt]
1, & m \in \mathcal{M}_{\mathrm{raw}}.
\end{cases}
\end{equation}

In parallel, we query the TKG for temporally valid facts and map them to their originating  chat turns via the bidirectional index:
\begin{equation}
\begin{aligned}
\mathcal{F}_T &= \{ (e_s,r,e_o,t)\in \mathcal{G} \mid t \in T_q \}, \\
\mathcal{S}_T &= \mathrm{Idx}(\mathcal{F}_T).
\end{aligned}
\end{equation}

where $\mathrm{Idx}(\cdot)$ returns the set of raw chat turns linked to the facts.

Finally, we rerank candidates through a composite scoring function that prioritizes semantic-time alignment before semantic similarity. To be specific, we use the indicator $\mathbb{I}[\tau(m)\in T_q]$ as the primary key and $s_{\mathrm{sem}}(m;q)$ as the secondary key:
\begin{equation}
\pi(m;q)=\Big(\mathbb{I}\big[\tau(m)\in T_q\big],\ s_{\mathrm{sem}}(m;q)\Big),
\end{equation}
and sort candidates in descending lexicographic order of $\pi(\cdot)$:
\begin{equation}
\mathcal{R}=\mathrm{Sort}_{m\in\mathcal{D}}\ \pi(m;q).
\end{equation}

This design enforces the query time constraint on compact summaries (topics/personas) while using TKG-grounded evidence that anchored on the Semantic timeline to promote relevant chat turns, yielding more accurate and contextually grounded retrieval.

\begin{table*}[htbp]
\centering
\label{tab:longmemeval_result}
\caption{Category-wise Accuracy For \textsc{LongMemEval\_S}. Accuracy (\%) by method across question types. Parentheses indicate category proportion and sample size. For the frontier-model variants, memory construction remains unchanged and uses GPT-4o-mini for graph extraction; only the answer generation model is replaced with the corresponding frontier model.}
\vspace{0.5ex}
\renewcommand{\arraystretch}{1.1}
\setlength{\tabcolsep}{3pt}
\small
\scalebox{0.95}{
\begin{threeparttable}
\begin{tabular}{lccccccc}
\toprule
\textbf{Method} & 
\textbf{ACC (\%)} &
\makecell{\textbf{Temporal}\\\makecell{\footnotesize ( $n{=}133$)}} &
\makecell{\textbf{Multi-Session}\\\makecell{\footnotesize ($n{=}133$)}} &
\makecell{\textbf{Knowledge-Update}\\\makecell{\footnotesize ($n{=}78$)}} &
\makecell{\textbf{Single-User}\\\makecell{\footnotesize ($n{=}70$)}} &
\makecell{\textbf{Single-Assistant}\\\makecell{\footnotesize ($n{=}56$)}} &
\makecell{\textbf{Single-Preference}\\\makecell{\footnotesize ($n{=}30$)}} \\
\midrule

\rowcolor{gptbar}\multicolumn{8}{c}{$\vcenter{\hbox{\openailogo}}$\ \textbf{GPT-4o-mini}}\\
\midrule
\rowcolor{gptgray} \textit{Full Text}  & 56.80 & 31.58 & 45.45 & \underline{76.92} & 87.14 & 89.29 & 36.67 \\
\rowcolor{gptgray} \textit{Naive RAG}  & 61.00 & 39.85 & 48.48 & 67.95 & {90.00} & \textbf{98.21} & 53.33 \\
\rowcolor{gptgray} \textit{LangMem}   & 37.20 & 15.79 & 20.30 & 66.67 & 60.00 & 46.43 & 60.00 \\
\rowcolor{gptgray} \textit{A-MEM}    & \underline{62.60} & \underline{47.36} & \underline{48.87} & 64.11 & \textbf{92.86} & {96.43} & 46.67 \\
\rowcolor{gptgray} \textit{Zep}    &  60.20 & 36.50 & 47.40 & 76.90 & 81.40 & 81.80 & 30.00 \\
\rowcolor{gptgray} \textit{MemoryOS}  & 44.80 & 32.33 & 31.06 & 48.72 & 80.00 & 64.29 & 30.00 \\
\rowcolor{gptgray} \textit{Mem0}    & 53.61 & 40.15 & 46.21 & 70.12 & 81.43 & 41.07 & \textbf{60.00} \\

\rowcolor{gptgray} \ours{}  & \textbf{74.80} & \textbf{69.92} & \textbf{69.17} & \textbf{80.77} & \underline{87.14} & \underline{94.64} & \underline{40.00} \\\midrule

\rowcolor{qwenbar}\multicolumn{8}{c}{$\vcenter{\hbox{\qwenlogo}}$\ \textbf{Qwen3-30B-A3B-Instruct-2507}}\\
\midrule
\rowcolor{qwenlav} \textit{Full Text}  & 54.80 & 33.08 & 35.61 & 76.92 & 82.86 & 87.50 & 50.00 \\
\rowcolor{qwenlav} \textit{Naive RAG}  & 60.80 & 36.84 & 47.73 & 65.38 & \underline{91.43} & \textbf{98.21} & 70.00 \\
\rowcolor{qwenlav} \textit{LangMem}   &  50.80 & 37.60 & 38.35 & 67.95 & 78.57 & 42.86 & \underline{70.00} \\
\rowcolor{qwenlav} \textit{A-MEM}    &  \underline{65.20} & \underline{51.88} & \underline{51.12} & \underline{76.93} & 90.00 & 96.43 & 40.00 \\
\rowcolor{qwenlav} \textit{MemoryOS}  &  49.60 & 28.57 & 36.84 & 61.54 & 72.86 & 92.86 & 33.33 \\
\rowcolor{qwenlav} \textit{Mem0}   &  39.51 & 41.94 & 28.13 & 28.57 & 55.32 & 26.09 & \textbf{81.82} \\
\rowcolor{qwenlav} \ours{}  &\textbf{74.80}  &\textbf{63.91} &\textbf{63.91} & \textbf{82.05} & \textbf{97.14} &  \underline{92.86} &66.67\\
\midrule
\rowcolor{gray!20}\multicolumn{8}{c}{\textbf{TSM with Frontier Models}}\\
\midrule
\rowcolor{gray!10}
\shortstack[l]{\ours{} +\\ \textit{Gemini-3-Pro}}
& 74.60 & 60.90 & 69.17 & \textbf{93.59} & 92.86 & \textbf{91.07} & 36.67 \\

\rowcolor{gray!10}
\shortstack[l]{\ours{} +\\ \textit{Gemini-3-Flash}}
& \textbf{79.60} & \textbf{71.43} & \textbf{75.19} & 92.31 & 92.86 & 89.29 & \textbf{53.33} \\

\bottomrule
\end{tabular}
\end{threeparttable}
}
\label{tab:catwise_acc_longmemeval}
\end{table*}

\subsection{Hierarchical Memory Update}
\label{sec:update}

Memory in \ours{} is maintained via a dual-stage update mechanism that separates low-latency graph maintenance from high-cost duration consolidation.

\noindent \textbf{Lightweight Online Graph Update.}
As new dialogue turns arrive, we update the TKG as the \emph{episodic memory} in an incremental manner, without blocking online inference. 
For each extracted candidate entity, \textit{add} creates a new node when the mention corresponds to an unseen entity; \textit{merge} maps the mention to an existing node and integrates newly observed attributes or contextual evidence to enrich the entity representation while avoiding duplication. 
Relations are updated in a temporally grounded way that each fact is associated with a \textit{valid\_time} and an \textit{invalid\_time}, indicating the interval during which it holds. 
Given a new extracted fact, we compare it with existing edges in both semantics and time, and apply one of four operations: \textit{DUPLICATE}
, \textit{ADD}
, \textit{INVALIDATE}
, and \textit{UPDATE}. 
These lightweight operations keep the index chronologically faithful and consistent as interactions evolve.

\noindent \textbf{Sleep-time Summary Consolidation.}
In contrast to the memory graph, topic/persona summaries are high-level durative memories and are expensive to refresh. 
We therefore update summaries periodically (e.g., once per month, aligned with the summary time granularity) or when the accumulated turns exceed a preset threshold. 
During consolidation, we reorganize all entity mentions via GMM-based clustering and re-summarize the resulting clusters into updated topic and persona snapshots. 
This ``sleep-time'' procedure reduces construction cost and latency while preserving coherence over long horizons.


\begin{table*}[htbp]
\centering
\caption{Category-wise accuracy on \textsc{LoCoMo}. Accuracy (\%) of different memory systems across question types. For frontier-model variants, the memory graph is still constructed by GPT-4o-mini, while the final response is generated by the corresponding frontier model. Parentheses indicate category proportion and sample size.}
\small
\vspace{0.5ex}
\renewcommand{\arraystretch}{1.1}
\setlength{\tabcolsep}{3pt}
\scalebox{0.95}{
\begin{threeparttable}
\begin{tabular}{lccccc}
\toprule
\textbf{Method} & 
\textbf{ACC (\%)} &
\makecell{\textbf{Temporal}\\{\footnotesize ($n{=}321$)}} &
\makecell{\textbf{Multi-Hop}\\{\footnotesize ($n{=}282$)}} &
\makecell{\textbf{Open-Domain}\\{\footnotesize ($n{=}96$)}} &
\makecell{\textbf{Single-Hop}\\{\footnotesize ($n{=}841$)}} \\
\midrule

\rowcolor{gptbar}\multicolumn{6}{c}{$\vcenter{\hbox{\openailogo}}$\ \textbf{GPT-4o-mini} }\\
\midrule
\rowcolor{gptgray} \textit{Full Text}   
& \underline{71.83} & \textbf{76.92} & 87.14 & 89.29 & 36.67 \\

\rowcolor{gptgray} \textit{Naive RAG}   
& 63.64 & 67.95 & \underline{90.00} & \textbf{98.21} & 53.33 \\

\rowcolor{gptgray} \textit{LangMem}    
& 57.20 & 66.67 & 60.00 & 46.43 & 60.00 \\

\rowcolor{gptgray} \textit{A-MEM}    
& 64.16 & 64.11 & \textbf{92.86} & \underline{96.43} & 46.67 \\

\rowcolor{gptgray} \textit{MemoryOS}   
& 58.25 & 48.72 & 80.00 & 64.29 & 30.00 \\

\rowcolor{gptgray} \textit{Mem0$^g$}    
& 68.44 & 58.13 & 47.19 & 75.71 & 65.71 \\

\rowcolor{gptgray} \textit{Zep}  
& 58.44 & 61.65 & 71.79 & 88.57 & \textbf{91.07} \\ 

\rowcolor{gptgray} \ours{}  
& \textbf{76.69} & \underline{71.03} & 66.67 & 58.33 & \underline{84.30} \\

\addlinespace[0.4ex]
\midrule

\rowcolor{qwenbar}\multicolumn{6}{c}{$\vcenter{\hbox{\qwenlogo}}$\ \textbf{Qwen3-30B-A3B-Instruct-2507}}\\
\midrule
\rowcolor{qwenlav} \textit{Full Text}   
& \textbf{74.87} & \textbf{76.92} & \underline{82.86} & \underline{87.50} & 50.00 \\

\rowcolor{qwenlav} \textit{Naive RAG}   
& 66.95 & 65.38 & \textbf{91.43} & \textbf{98.21} & 70.00 \\

\rowcolor{qwenlav} \textit{LangMem}    
& 60.53 & 67.95 & 78.57 & 42.86 & 70.00 \\

\rowcolor{qwenlav} \textit{A-MEM}     
& 56.10 & \underline{76.93} & \underline{90.00} & \underline{96.43} & 40.00 \\

\rowcolor{qwenlav} \textit{MemoryOS}   
& 61.04 & 61.54 & 72.86 & 92.86 & 33.33 \\

\rowcolor{qwenlav} \textit{Mem0}   
& 43.31 & 28.57 & 55.32 & 26.09 & \textbf{81.82} \\

\rowcolor{qwenlav} \ours{} 
& \underline{71.23} & 65.42 & 64.54 & 56.25 & \underline{77.41} \\
\addlinespace[0.4ex]
\midrule
\rowcolor{gray!20}\multicolumn{6}{c}{\textbf{TSM with Frontier Models}}\\
\midrule
\rowcolor{gray!10}
\shortstack[l]{\ours{} +\\ \textit{Gemini-3-Pro}}
& 66.30 & 63.86 & 50.71 & 46.88 & 74.67 \\

\rowcolor{gray!10}
\shortstack[l]{\ours{} +\\ \textit{Gemini-3-Flash}}
& 73.96 & \textbf{77.26} & 57.45 & 51.04 & 80.86 \\
\bottomrule
\end{tabular}
\end{threeparttable}
}
\label{tab:catwise_acc_by_locomo}
\end{table*}

\section{Experiments}

In this section, we evaluate \ours{} on real-world datasets to assess its performance. 
\subsection{Experimental Setup}

\noindent \textbf{Dataset.} The performance of \ours{} was evaluated on two public long-term memory benchmarks: \textsc{LongMemEval} and \textsc{LoCoMo}.
\textsc{LongMemEval}~\cite{wu_longmemeval_2025} is a comprehensive benchmark for assessing the long-term memory capabilities of chat assistants. It consists of 500 manually created questions to test five core memory abilities: information extraction, multi-session reasoning, temporal reasoning, knowledge updates, and abstention. Each question requires recalling information hidden within one or more task-oriented dialogues between a user and an assistant.
We utilize \textsc{\longmemeval\_S}, a version where each question has approximately 115k tokens as its history. 
\textsc{LoCoMo}~\cite{maharana_evaluating_2024} focuses on extremely long multi-session dialogues, containing 1,986 questions in five distinct categories: single-hop, multi-hop, temporal, open-domain, and adversarial reasoning.


\noindent \textbf{Evaluation Metrics.}
We report \textbf{Accuracy (ACC)} for effectiveness, defined as the proportion of correctly answered questions. 
Following prior work, evaluation is conducted with \emph{GPT-4.1-mini} as an LLM judge, guided by a detailed evaluation prompt (see Appendix~\ref{prompts:LLM-as-Judge}). 

\noindent \textbf{Baselines.} We compare \ours{} against several representative baselines of conversational memory
modeling. (1) Full Text, (2) Naive RAG, (3) LangMem~\cite{langchain_langchain_nodate}, (4) A-MEM~\cite{xu2025mem}, (5) MemoryOS~\citep{kang_memory_nodate}, 
(6) Mem0 or Mem0$^g$ (a graph variant of Mem0)~\citep{chhikara2025mem0}, (7) Zep~\citep{rasmussen_zep_2025}. In addition, all methods use GPT-4o-mini~\cite{openai_gpt-4o_2024} and Qwen3-30B-A3B-Instruct-2507~\cite{qwen3technicalreport} as the LLM backbones. 


\subsection{Main Results}

Tables \ref{tab:catwise_acc_longmemeval} and \ref{tab:catwise_acc_by_locomo} report the category-wise performance on \textsc{LongMemEval\_S} and \textsc{LoCoMo}, respectively. Across both datasets, our method consistently achieves strong performance, outpacing existing baselines and demonstrating robust memory reasoning capabilities in diverse long-context scenarios.

\noindent{\textbf{\textsc{LongMemEval}.}} On \textsc{LongMemEval\_S}, \ours{} achieves the highest overall accuracy of 74.80\%, surpassing the A-MEM baseline (62.60\%) on GPT-4o-mini. We set new state-of-the-art results on \textit{Temporal}, \textit{Multi-Session}, and \textit{Knowledge-Update} questions, all of which are strongly dependent on time. The notable improvement in \textit{Multi-Session} accuracy (+20.30\%) underscores the crucial role of durative memory. By maintaining long-term contextual information, \ours{} enables more coherent and accurate reasoning across multiple interactions. The improvement in \textit{Temporal} accuracy (+22.56\%) highlights the effectiveness of using a semantic timeline. By leveraging temporal semantics, \ours{} can retrieve and update context across time, leading to more accurate handling of time-sensitive queries.
Performance on \textit{Single-Session Preference} questions is lower; however, the limited sample size and high variance suggest that this category does not significantly impact overall performance.
\ours{} also achieves superior performance on Qwen3 backbones, with an accuracy of 74.80\%, and yields the best results on \textit{Single-User} questions, showing strong performance across all categories.

\noindent{\textbf{\textsc{LoCoMo}.}} On \textsc{LoCoMo}, the full-text baseline achieves the highest performance on Qwen3-30B-A3B-Instruct-2507. This is expected because \textsc{LoCoMo} conversations are relatively short (16k to 26k tokens) compared to \textsc{LongMemEval\_S} (115k tokens), making it feasible to fit the entire context within the model's window. More importantly, \textsc{LoCoMo} does not effectively test critical memory capabilities such as knowledge updates. When the full conversation is short enough to process directly, providing complete context naturally outperforms any retrieval-based approach that risks information loss.

Nevertheless, among all memory-based methods, \ours{} achieves the best performance with 71.23\% accuracy on Qwen3-30B and 76.69\% on GPT-4o-mini, substantially outperforming Naive RAG (63.64\%) and Mem0$^g$ (68.44\%). Our method excels particularly on \textit{Single-Hop} and \textit{Temporal} questions, demonstrating that our temporal grounding and hierarchical memory organization provide superior retrieval precision compared to existing memory systems.


\begin{table*}[t]
\centering
\setlength{\tabcolsep}{5pt}
\small
\caption{Ablation results on \textsc{LongMemEval\_S}. Best numbers in each column are in bold. $\Delta$ denotes the absolute change in accuracy points, and Rel.\% denotes the relative change with respect to \ours{}.}
\label{tab:ablation_horizontal}
\begin{tabular}{lccccccc}
\toprule
\thead{Name} &
\thead{Overall} &
\thead{Single-Session\\User} &
\thead{Temporal} &
\thead{Knowledge\\Update} &
\thead{Multi-\\Session} &
\thead{Single-Session\\Preference} &
\thead{Single-Session\\Assistant} \\
\midrule

\rowcolor{gray!10}
\textbf{\ours{}} 
& \textbf{74.80} 
& 87.14 
& \textbf{69.92} 
& 80.77 
& 69.17 
& \textbf{40.00} 
& 94.64 \\

\cmidrule(lr){1-8}

w/o Temporal 
& 72.80 
& 87.14 
& 63.91 
& 79.49 
& \textbf{71.43} 
& 33.33 
& 91.07 \\
$\Delta$     
& $-2.0$ 
& $+0.0$ 
& $-6.0$ 
& $-1.3$ 
& $+2.3$ 
& $-6.7$ 
& $-3.6$ \\
Rel.\%       
& $\downarrow 2.7\%$ 
& $0.0\%$ 
& $\downarrow 8.6\%$ 
& $\downarrow 1.6\%$ 
& $\uparrow 3.3\%$ 
& $\downarrow 16.7\%$ 
& $\downarrow 3.8\%$ \\

\cmidrule(lr){1-8}

w/o Persona/Summary 
& 73.40 
& \textbf{88.57} 
& 65.41 
& \textbf{82.05} 
& 69.92 
& 23.33 
& \textbf{96.43} \\
$\Delta$     
& $-1.4$ 
& $+1.4$ 
& $-4.5$ 
& $+1.3$ 
& $+0.8$ 
& $-16.7$ 
& $+1.8$ \\
Rel.\%       
& $\downarrow 1.9\%$ 
& $\uparrow 1.6\%$ 
& $\downarrow 6.5\%$ 
& $\uparrow 1.6\%$ 
& $\uparrow 1.1\%$ 
& $\downarrow 41.7\%$ 
& $\uparrow 1.9\%$ \\

\bottomrule
\end{tabular}
\end{table*}

\subsection{Efficiency}
Table R2 reports both token cost and recall latency on \textsc{LongMemEval\_S}. Although \ours{} does not use the fewest tokens overall, it achieves the best task performance (74.80\% ACC). It also delivers the fastest recall latency, with a P50/P95 of 1.57/2.39 seconds. 

These results suggest that the efficiency advantage of \ours{} mainly comes from its online retrieval design, rather than minimizing total token usage. Specifically, expensive memory construction is shifted to the update stage. During inference, the model only performs temporal filtering over a pre-built memory graph and local reranking over retrieved candidates. This avoids additional LLM-based extraction during recall. Moreover, the sleep-time update variant further reduces the total token cost from 2065.54k to 1959.53k. This shows that periodic offline summarization can reduce update overhead without affecting online latency. Overall, the results indicate that \ours{} offers a strong trade-off between effectiveness and practical inference efficiency for long-horizon memory retrieval.

\begin{table*}[t]
\centering
\caption{Efficiency comparison on \textsc{LongMemEval\_S}. We report accuracy, token cost (in thousands), and recall latency in seconds. Lower latency is better.}
\label{tab:efficiency_longmemevals}
\vspace{0.5ex}
\renewcommand{\arraystretch}{1.12}
\setlength{\tabcolsep}{4pt}
\small
\scalebox{0.95}{
\begin{threeparttable}
\begin{tabular}{lccccccc}
\toprule
\multirow{2}{*}{\textbf{Method}} 
& \multirow{2}{*}{\textbf{ACC}} 
& \multicolumn{5}{c}{\textbf{Token Cost (k)}} 
& \multirow{2}{*}{\makecell{\textbf{Recall Latency} $\downarrow$\\\textbf{(P50 / P95, s)}}} \\
\cmidrule(lr){3-7}
& & \textbf{Summary In} & \textbf{Summary Out} & \textbf{Update In} & \textbf{Update Out} & \textbf{Total} & \\
\midrule

\rowcolor{gray!20}\multicolumn{8}{c}{\textbf{Baseline Methods}}\\
\midrule
\rowcolor{gray!10} A-MEM    & 62.60 & 214.66   & 42.82  & 1,157.52 & 190.81   & 1,605.81 & 5.12 / 11.89 \\
\rowcolor{gray!10} Mem0     & 53.61 & 424.13   & 17.76  & 150.56   & 1,152.62 & 1,745.07 & 3.64 / 6.11  \\
\rowcolor{gray!10} MemoryOS & 44.80 & 2,302.35 & 304.18 & 350.02   & 35.19    & 2,991.74 & 1.63 / 3.95  \\
\midrule

\rowcolor{gptbar}\multicolumn{8}{c}{\textbf{TSM Variants}}\\
\midrule
\rowcolor{gptgray} \textbf{TSM (Ours)} & \textbf{74.80} & 1,430.71 & 22.09 & 609.09 & 3.65 & 2,065.54 & \textbf{1.57 / 2.39} \\
\rowcolor{gptgray} \textbf{TSM (Sleep-Time Update)} & -- & 1,334.06 & 12.72 & -- & -- & 1,959.53 & -- \\

\bottomrule
\end{tabular}
\end{threeparttable}
}
\end{table*}

\subsection{Ablation Study}
To evaluate the effectiveness of \ours{}, we perform an ablation study on \textsc{LongMemEval\_S} under two settings. For the semantic timeline, the ``w.o. temporal'' variant removes temporal ranking and filtering during memory utilization and relies solely on dense retrieval. As a result, durative summaries cannot be selected based on the query's temporal intent, and semantically and temporally related chat turns are not prioritized. For durative memories, ``w.o. summary'' removes all topics and personas while retaining temporal reranking, ensuring that this configuration is not RAG-based.

As shown in Table~\ref{tab:ablation_horizontal}, removing temporal retrieval (``w.o.temporal'') leads to a noticeable degradation in overall performance (74.8\% $\rightarrow$ 72.8\%, $-2.0$), with the largest drop observed in \textit{Temporal} questions ($-6.0$). This highlights the importance of explicit temporal modeling for addressing time-related queries. It also negatively impacts \textit{Single-Session-Assistant} queries ($-3.5$) and \textit{Single-Session Preference} queries ($-6.7$), indicating that temporal misalignment affects downstream response generation even when the query is not explicitly time-sensitive.

Removing summaries while keeping temporal reranking (``w.o.summary'') results in a decrease in overall accuracy (74.8\% $\rightarrow$ 73.6\%, $-1.2$), and notably hurts \textit{Single-Session-Preference} ($-10.0$). This suggests that persona information can be beneficial for user-specific and preference-centric queries. 
It also significantly harms \textit{Temporal} tasks ($-4.5$), implying that durative memories are crucial for capturing the narrative of past experiences.
However, it improves performance on \textit{Single-Session-User} (+1.5), \textit{Knowledge-Update} (+1.3), and \textit{Multi-Session} (+1.5), possibly due to the removal of distractions in certain categories.

Overall, both components contribute to performance, with temporal modeling offering the most consistent and substantial improvements.

\section{Conclusions}

In this paper, we presented \ours{}, a novel approach for memory construction and utilization that addresses key limitations of existing methods. By incorporating both semantic-time grounded and duration-aware management, \ours{} overcame the fragmentation caused by isolated, point-wise memory entries, ensuring that long-term, continuous user experiences are captured. The proposed duration-aware memory construction consolidates temporally continuous and semantically related information, enhancing contextual consistency and enabling more accurate retrieval of relevant memories. Additionally, the integration of semantic-time guided memory utilization improved retrieval by considering the temporal intent behind user queries. Extensive experiments on \textsc{LongMemEval} and \textsc{LoCoMo} datasets demonstrated that \ours{} significantly outperforms existing memory baselines, achieving notable improvements in QA accuracy, particularly in tasks requiring multi-session understanding and temporal reasoning. These results highlighted the importance of modeling semantic time and duration for effective, reliable long-term memory in LLM-based agents.

\section*{Limitations}
While \ours{} demonstrates significant improvements in retrieval relevance and response quality, several limitations warrant discussion. 
First, 
\ours{} adopts a fixed temporal granularity (e.g., monthly intervals) for grouping durative summaries, which may not be optimal across all application domains. Adaptive granularity selection based on the temporal density of events could improve flexibility but is left for future work.
Second, our work focuses on personalization applications. Extending our approach to other memory paradigms, such as procedural memory for agent learning and shared memory for multi-agent systems, remains important future work.

\section*{Acknowledgments}
This work is partially funded by the National Key Research and Development Program of China under Grants No. 2024YFC3308200, the Strategic Priority Research Program of the CAS under Grants No. XDB0680102, the National Natural Science Foundation of China under grants 62306299, 62441229 and 62406308. We thank anonymous reviewers for their insightful comments and suggestions.

\bibliography{main}

@article{yang2026graph,
  title={Graph-based Agent Memory: Taxonomy, Techniques, and Applications},
  author={Yang, Chang and Zhou, Chuang and Xiao, Yilin and Dong, Su and Zhuang, Luyao and Zhang, Yujing and Wang, Zhu and Hong, Zijin and Yuan, Zheng and Xiang, Zhishang and others},
  journal={arXiv preprint arXiv:2602.05665},
  year={2026}

}

@online{wuGAMHierarchicalGraphbased2026a,
  title = {{{GAM}}: {{Hierarchical Graph-based Agentic Memory}} for {{LLM Agents}}},
  shorttitle = {{{GAM}}},
  author = {Wu, Zhaofen and Zhang, Hanrong and Lin, Fulin and Xu, Wujiang and Xu, Xinran and Chen, Yankai and Zou, Henry Peng and Chen, Shaowen and Zhang, Weizhi and Liu, Xue and Yu, Philip S. and Wang, Hongwei},
  date = {2026-04-14},
  eprint = {2604.12285},
  eprinttype = {arXiv},
  eprintclass = {cs},
  doi = {10.48550/arXiv.2604.12285},
  url = {http://arxiv.org/abs/2604.12285},
  urldate = {2026-04-15},
  pubstate = {prepublished}
}

@misc{zhang2026expseek,
      title={ExpSeek: Self-Triggered Experience Seeking for Web Agents},
      author={Wenyuan Zhang and Xinghua Zhang and Haiyang Yu and Shuaiyi Nie and Bingli Wu and Juwei Yue and Tingwen Liu and Yongbin Li},
      year={2026},
      eprint={2601.08605},
      archivePrefix={arXiv},
      primaryClass={cs.CL},
      url={https://arxiv.org/abs/2601.08605},
}

@article{wu2026knowme,
  title={KnowMe-Bench: Benchmarking Person Understanding for Lifelong Digital Companions},
  author={Wu, Tingyu and Chen, Zhisheng and Weng, Ziyan and Wang, Shuhe and Li, Chenglong and Zhang, Shuo and Hu, Sen and Wu, Silin and Lan, Qizhen and Wang, Huacan and others},
  journal={arXiv preprint arXiv:2601.04745},
  year={2026}
}

@misc{qwen3technicalreport,
      title={Qwen3 Technical Report}, 
      author={{Qwen Team}},
      year={2025},
      eprint={2505.09388},
      archivePrefix={arXiv},
      primaryClass={cs.CL},
      url={https://arxiv.org/abs/2505.09388}, 
}

@article{Honnibal2020spaCy,
  title         = {spaCy: Industrial-strength Natural Language Processing in Python},
  author        = {Matthew Honnibal and Ines Montani and Sofie Van Landeghem and Adriane Boyd},
  year          = {2020},
  doi           = {10.5281/zenodo.1212303}
}

@misc{owl2025,
  title={OWL: Optimized Workforce Learning for General Multi-Agent Assistance in Real-World Task Automation},
  author={Camel-AI},
  url={https://github.com/camel-ai/owl},
  year={2025}
}

@misc{openmanus2025,
  author = {Xinbin Liang and Jinyu Xiang and Zhaoyang Yu and Jiayi Zhang and Sirui Hong and Sheng Fan and Xiao Tang},
  title = {OpenManus: An open-source framework for building general AI agents},
  year = {2025},
  publisher = {Zenodo},
  doi = {10.5281/zenodo.15186407},
  url = {https://doi.org/10.5281/zenodo.15186407},
}

@misc{google2025deepresearch,
  author       = {{Google}},
  title        = {Gemini Deep Research — your personal research assistant},
  year         = {2025},
  howpublished = {\url{https://gemini.google/overview/deep-research/?hl=en-GB}}
}

@misc{bytedance2025deerflow,
  author       = {{ByteDance}},
  title        = {DeerFlow: Deep Exploration and Efficient Research Framework},
  year         = {2025},
  howpublished = {\url{https://deerflow.tech/z}}
}

@article{shinn2023reflexion,
  title={Reflexion: Language agents with verbal reinforcement learning},
  author={Shinn, Noah and Cassano, Federico and Gopinath, Ashwin and Narasimhan, Karthik and Yao, Shunyu},
  journal={Advances in Neural Information Processing Systems},
  volume={36},
  pages={8634--8652},
  year={2023}
}

@article{sumers2023cognitive,
  title={Cognitive architectures for language agents},
  author={Sumers, Theodore and Yao, Shunyu and Narasimhan, Karthik and Griffiths, Thomas},
  journal={Transactions on Machine Learning Research},
  year={2023}
}

@article{chhikara2025mem0,
  title={Mem0: Building production-ready ai agents with scalable long-term memory},
  author={Chhikara, Prateek and Khant, Dev and Aryan, Saket and Singh, Taranjeet and Yadav, Deshraj},
  journal={arXiv preprint arXiv:2504.19413},
  year={2025}
}

@article{xu2025mem,
  title={A-mem: Agentic memory for llm agents},
  author={Xu, Wujiang and Mei, Kai and Gao, Hang and Tan, Juntao and Liang, Zujie and Zhang, Yongfeng},
  journal={arXiv preprint arXiv:2502.12110},
  year={2025}
}

@misc{openai_gpt-4o_2024,
	title = {{GPT}-4o mini: advancing cost-efficient intelligence},
	shorttitle = {{GPT}-4o mini},
	url = {https://openai.com/index/gpt-4o-mini-advancing-cost-efficient-intelligence/},
	language = {en-US},
	urldate = {2026-01-06},
	author = {{OpenAI}},
	month = jul,
	year = {2024},
}

@book{kadavy_digital_2021,
	title = {Digital {Zettelkasten}: {Principles}, {Methods}, \& {Examples}},
	shorttitle = {Digital {Zettelkasten}},
	language = {en},
	publisher = {Kadavy, Inc.},
	author = {Kadavy, David},
	month = may,
	year = {2021},
	note = {Google-Books-ID: o4gwEAAAQBAJ},
}

@article{macdonald_hippocampal_2011,
	title = {Hippocampal "time cells" bridge the gap in memory for discontiguous events},
	volume = {71},
	issn = {1097-4199},
	doi = {10.1016/j.neuron.2011.07.012},
	language = {eng},
	number = {4},
	journal = {Neuron},
	author = {MacDonald, Christopher J. and Lepage, Kyle Q. and Eden, Uri T. and Eichenbaum, Howard},
	month = aug,
	year = {2011},
	pmid = {21867888},
	pmcid = {PMC3163062},
	note = {TLDR: A robust hippocampal representation of sequence memories is reported, highlighted by "time cells" that encode successive moments during an empty temporal gap between the key events, while also encoding location and ongoing behavior.},
	pages = {737--749},
}

@misc{huet_episodic_2025,
	title = {Episodic {Memories} {Generation} and {Evaluation} {Benchmark} for {Large} {Language} {Models}},
	url = {http://arxiv.org/abs/2501.13121},
	doi = {10.48550/arXiv.2501.13121},
	urldate = {2026-01-02},
	publisher = {arXiv},
	author = {Huet, Alexis and Houidi, Zied Ben and Rossi, Dario},
	month = jan,
	year = {2025},
	note = {arXiv:2501.13121 [cs]
TLDR: It is argued that integrating episodic memory capabilities into LLM is essential for advancing AI towards human-like cognition, increasing their potential to reason consistently and ground their output in real-world episodic events, hence avoiding confabulations.},
}

@misc{hu_memory_2025,
	title = {Memory in the {Age} of {AI} {Agents}},
	url = {http://arxiv.org/abs/2512.13564},
	doi = {10.48550/arXiv.2512.13564},
	urldate = {2025-12-31},
	publisher = {arXiv},
	author = {Hu, Yuyang and Liu, Shichun and Yue, Yanwei and Zhang, Guibin and Liu, Boyang and Zhu, Fangyi and Lin, Jiahang and Guo, Honglin and Dou, Shihan and Xi, Zhiheng and Jin, Senjie and Tan, Jiejun and Yin, Yanbin and Liu, Jiongnan and Zhang, Zeyu and Sun, Zhongxiang and Zhu, Yutao and Sun, Hao and Peng, Boci and Cheng, Zhenrong and Fan, Xuanbo and Guo, Jiaxin and Yu, Xinlei and Zhou, Zhenhong and Hu, Zewen and Huo, Jiahao and Wang, Junhao and Niu, Yuwei and Wang, Yu and Yin, Zhenfei and Hu, Xiaobin and Liao, Yue and Li, Qiankun and Wang, Kun and Zhou, Wangchunshu and Liu, Yixin and Cheng, Dawei and Zhang, Qi and Gui, Tao and Pan, Shirui and Zhang, Yan and Torr, Philip and Dou, Zhicheng and Wen, Ji-Rong and Huang, Xuanjing and Jiang, Yu-Gang and Yan, Shuicheng},
	month = dec,
	year = {2025},
	note = {arXiv:2512.13564 [cs]},
}

@misc{fang_lightmem_2025,
	title = {{LightMem}: {Lightweight} and {Efficient} {Memory}-{Augmented} {Generation}},
	shorttitle = {{LightMem}},
	url = {http://arxiv.org/abs/2510.18866},
	doi = {10.48550/arXiv.2510.18866},
	urldate = {2025-12-29},
	publisher = {arXiv},
	author = {Fang, Jizhan and Deng, Xinle and Xu, Haoming and Jiang, Ziyan and Tang, Yuqi and Xu, Ziwen and Deng, Shumin and Yao, Yunzhi and Wang, Mengru and Qiao, Shuofei and Chen, Huajun and Zhang, Ningyu},
	month = nov,
	year = {2025},
	note = {arXiv:2510.18866 [cs]
TLDR: Inspired by the Atkinson-Shiffrin model of human memory, LightMem organizes memory into three complementary stages, which strikes a balance between the performance and efficiency of memory systems.},
}

@misc{langchain_langchain_nodate,
	title = {{LangChain} {Blog}},
	url = {https://blog.langchain.com/},
	urldate = {2025-12-29},
	author = {{LangChain}},
}

@misc{huang_retrieval-augmented_2025,
	title = {Retrieval-{Augmented} {Generation} with {Hierarchical} {Knowledge}},
	url = {http://arxiv.org/abs/2503.10150},
	doi = {10.48550/arXiv.2503.10150},
	urldate = {2025-12-29},
	publisher = {arXiv},
	author = {Huang, Haoyu and Huang, Yongfeng and Yang, Junjie and Pan, Zhenyu and Chen, Yongqiang and Ma, Kaili and Chen, Hongzhi and Cheng, James},
	month = sep,
	year = {2025},
	note = {arXiv:2503.10150 [cs]
TLDR: This paper introduces a new RAG approach, called HiRAG, which utilizes hierarchical knowledge to enhance the semantic understanding and structure capturing capabilities of RAG systems in the indexing and retrieval processes.},
}

@misc{zhang_g-memory_2025,
	title = {G-{Memory}: {Tracing} {Hierarchical} {Memory} for {Multi}-{Agent} {Systems}},
	shorttitle = {G-{Memory}},
	url = {http://arxiv.org/abs/2506.07398},
	doi = {10.48550/arXiv.2506.07398},
	urldate = {2025-12-29},
	publisher = {arXiv},
	author = {Zhang, Guibin and Fu, Muxin and Wan, Guancheng and Yu, Miao and Wang, Kun and Yan, Shuicheng},
	month = jun,
	year = {2025},
	note = {arXiv:2506.07398 [cs]
TLDR: G-Memory is introduced, a hierarchical, agentic memory system for MAS inspired by organizational memory theory, which manages the lengthy MAS interaction via a three-tier graph hierarchy: insight, query, and interaction graphs.},
}

@misc{li_hello_2025,
	title = {Hello {Again}! {LLM}-powered {Personalized} {Agent} for {Long}-term {Dialogue}},
	url = {http://arxiv.org/abs/2406.05925},
	doi = {10.48550/arXiv.2406.05925},
	urldate = {2025-12-29},
	publisher = {arXiv},
	author = {Li, Hao and Yang, Chenghao and Zhang, An and Deng, Yang and Wang, Xiang and Chua, Tat-Seng},
	month = feb,
	year = {2025},
	note = {arXiv:2406.05925 [cs]},
}

@misc{wu_human_2025,
	title = {From {Human} {Memory} to {AI} {Memory}: {A} {Survey} on {Memory} {Mechanisms} in the {Era} of {LLMs}},
	shorttitle = {From {Human} {Memory} to {AI} {Memory}},
	url = {http://arxiv.org/abs/2504.15965},
	doi = {10.48550/arXiv.2504.15965},
	urldate = {2025-12-29},
	publisher = {arXiv},
	author = {Wu, Yaxiong and Liang, Sheng and Zhang, Chen and Wang, Yichao and Zhang, Yongyue and Guo, Huifeng and Tang, Ruiming and Liu, Yong},
	month = apr,
	year = {2025},
	note = {arXiv:2504.15965 [cs]},
}

@misc{luo_large_2025,
	title = {Large {Language} {Model} {Agent}: {A} {Survey} on {Methodology}, {Applications} and {Challenges}},
	shorttitle = {Large {Language} {Model} {Agent}},
	url = {http://arxiv.org/abs/2503.21460},
	doi = {10.48550/arXiv.2503.21460},
	urldate = {2025-12-29},
	publisher = {arXiv},
	author = {Luo, Junyu and Zhang, Weizhi and Yuan, Ye and Zhao, Yusheng and Yang, Junwei and Gu, Yiyang and Wu, Bohan and Chen, Binqi and Qiao, Ziyue and Long, Qingqing and Tu, Rongcheng and Luo, Xiao and Ju, Wei and Xiao, Zhiping and Wang, Yifan and Xiao, Meng and Liu, Chenwu and Yuan, Jingyang and Zhang, Shichang and Jin, Yiqiao and Zhang, Fan and Wu, Xian and Zhao, Hanqing and Tao, Dacheng and Yu, Philip S. and Zhang, Ming},
	month = mar,
	year = {2025},
	note = {arXiv:2503.21460 [cs]},
}

@misc{minaee_large_2025,
	title = {Large {Language} {Models}: {A} {Survey}},
	shorttitle = {Large {Language} {Models}},
	url = {http://arxiv.org/abs/2402.06196},
	doi = {10.48550/arXiv.2402.06196},
	urldate = {2025-12-29},
	publisher = {arXiv},
	author = {Minaee, Shervin and Mikolov, Tomas and Nikzad, Narjes and Chenaghlu, Meysam and Socher, Richard and Amatriain, Xavier and Gao, Jianfeng},
	month = mar,
	year = {2025},
	note = {arXiv:2402.06196 [cs]},
}

@misc{matarazzo_survey_2025,
	title = {A {Survey} on {Large} {Language} {Models} with some {Insights} on their {Capabilities} and {Limitations}},
	url = {http://arxiv.org/abs/2501.04040},
	doi = {10.48550/arXiv.2501.04040},
	urldate = {2025-12-29},
	publisher = {arXiv},
	author = {Matarazzo, Andrea and Torlone, Riccardo},
	month = feb,
	year = {2025},
	note = {arXiv:2501.04040 [cs]},
}

@misc{sun_hierarchical_2025,
	title = {Hierarchical {Memory} for {High}-{Efficiency} {Long}-{Term} {Reasoning} in {LLM} {Agents}},
	url = {http://arxiv.org/abs/2507.22925},
	doi = {10.48550/arXiv.2507.22925},
	urldate = {2025-08-16},
	publisher = {arXiv},
	author = {Sun, Haoran and Zeng, Shaoning},
	month = jul,
	year = {2025},
	note = {arXiv:2507.22925 [cs]
TLDR: A Hierarchical Memory (H-MEM) architecture for LLM Agents is proposed that organizes and updates memory in a multi-level fashion based on the degree of semantic abstraction and consistently outperforms five baseline methods in long-term dialogue scenarios.},
}

@misc{nan_nemori_2025,
	title = {Nemori: {Self}-{Organizing} {Agent} {Memory} {Inspired} by {Cognitive} {Science}},
	shorttitle = {Nemori},
	url = {http://arxiv.org/abs/2508.03341},
	doi = {10.48550/arXiv.2508.03341},
	urldate = {2025-08-16},
	publisher = {arXiv},
	author = {Nan, Jiayan and Ma, Wenquan and Wu, Wenlong and Chen, Yize},
	month = aug,
	year = {2025},
	note = {arXiv:2508.03341 [cs]
TLDR: Nemori is a novel self-organizing memory architecture inspired by human cognitive principles that significantly outperforms prior state-of-the-art systems, with its advantage being particularly pronounced in longer contexts.},
}

@misc{yan_memory-r1_2025,
	title = {Memory-{R1}: {Enhancing} {Large} {Language} {Model} {Agents} to {Manage} and {Utilize} {Memories} via {Reinforcement} {Learning}},
	shorttitle = {Memory-{R1}},
	url = {http://arxiv.org/abs/2508.19828},
	doi = {10.48550/arXiv.2508.19828},
	urldate = {2025-09-02},
	publisher = {arXiv},
	author = {Yan, Sikuan and Yang, Xiufeng and Huang, Zuchao and Nie, Ercong and Ding, Zifeng and Li, Zonggen and Ma, Xiaowen and Schütze, Hinrich and Tresp, Volker and Ma, Yunpu},
	month = aug,
	year = {2025},
	note = {arXiv:2508.19828 [cs]
TLDR: Memory-R1 is presented, a reinforcement learning (RL) framework that equips LLMs with the ability to actively manage and utilize external memory through two specialized agents: a Memory Manager that learns structured operations, including ADD, UPDATE, DELETE, and NOOP and an Answer Agent that pre-selects and reasons over relevant entries.},
}

@misc{kim_pre-storage_2025,
	title = {Pre-{Storage} {Reasoning} for {Episodic} {Memory}: {Shifting} {Inference} {Burden} to {Memory} for {Personalized} {Dialogue}},
	shorttitle = {Pre-{Storage} {Reasoning} for {Episodic} {Memory}},
	url = {http://arxiv.org/abs/2509.10852},
	doi = {10.48550/arXiv.2509.10852},
	urldate = {2025-09-23},
	publisher = {arXiv},
	author = {Kim, Sangyeop and Lee, Yohan and Kim, Sanghwa and Kim, Hyunjong and Cho, Sungzoon},
	month = sep,
	year = {2025},
	note = {arXiv:2509.10852 [cs]
TLDR: This work introduces PREMem (Pre-storage Reasoning for Episodic Memory), a novel approach that shifts complex reasoning processes from inference to memory construction, and creates enriched representations while reducing computational demands during interactions.},
}

@misc{zhang_memgen_2025,
	title = {{MemGen}: {Weaving} {Generative} {Latent} {Memory} for {Self}-{Evolving} {Agents}},
	shorttitle = {{MemGen}},
	url = {http://arxiv.org/abs/2509.24704},
	doi = {10.48550/arXiv.2509.24704},
	urldate = {2025-10-01},
	publisher = {arXiv},
	author = {Zhang, Guibin and Fu, Muxin and Yan, Shuicheng},
	month = sep,
	year = {2025},
	note = {arXiv:2509.24704 [cs]
TLDR: MemGen is proposed, a dynamic generative memory framework that equips agents with a human-esque cognitive faculty and spontaneously evolves distinct human-like memory faculties, including planning memory, procedural memory, and working memory, suggesting an emergent trajectory toward more naturalistic forms of machine cognition.},
}

@misc{ouyang_reasoningbank_2025,
	title = {{ReasoningBank}: {Scaling} {Agent} {Self}-{Evolving} with {Reasoning} {Memory}},
	shorttitle = {{ReasoningBank}},
	url = {http://arxiv.org/abs/2509.25140},
	doi = {10.48550/arXiv.2509.25140},
	urldate = {2025-10-09},
	publisher = {arXiv},
	author = {Ouyang, Siru and Yan, Jun and Hsu, I.-Hung and Chen, Yanfei and Jiang, Ke and Wang, Zifeng and Han, Rujun and Le, Long T. and Daruki, Samira and Tang, Xiangru and Tirumalashetty, Vishy and Lee, George and Rofouei, Mahsan and Lin, Hangfei and Han, Jiawei and Lee, Chen-Yu and Pfister, Tomas},
	month = sep,
	year = {2025},
	note = {arXiv:2509.25140 [cs]
TLDR: Memory-driven experience scaling is established as a new scaling dimension, enabling agents to self-evolve with emergent behaviors naturally arise, and further introduces memory-aware test-time scaling (MaTTS), which accelerates and diversifies this learning process by scaling up the agent's interaction experience.},
}

@misc{kwon_embodied_2025,
	title = {Embodied {Agents} {Meet} {Personalization}: {Exploring} {Memory} {Utilization} for {Personalized} {Assistance}},
	shorttitle = {Embodied {Agents} {Meet} {Personalization}},
	url = {http://arxiv.org/abs/2505.16348},
	doi = {10.48550/arXiv.2505.16348},
	urldate = {2025-08-16},
	publisher = {arXiv},
	author = {Kwon, Taeyoon and Choi, Dongwook and Kim, Sunghwan and Kim, Hyojun and Moon, Seungjun and Kwak, Beong-woo and Huang, Kuan-Hao and Yeo, Jinyoung},
	month = may,
	year = {2025},
	note = {arXiv:2505.16348 [cs]
TLDR: MEMENTO is presented, a personalized embodied agent evaluation framework designed to comprehensively assess memory utilization capabilities to provide personalized assistance, which consists of a two-stage memory evaluation process design that enables quantifying the impact of memory utilization on task performance.},
}

@misc{zhou_mem1_2025,
	title = {{MEM1}: {Learning} to {Synergize} {Memory} and {Reasoning} for {Efficient} {Long}-{Horizon} {Agents}},
	shorttitle = {{MEM1}},
	url = {http://arxiv.org/abs/2506.15841},
	doi = {10.48550/arXiv.2506.15841},
	urldate = {2025-08-16},
	publisher = {arXiv},
	author = {Zhou, Zijian and Qu, Ao and Wu, Zhaoxuan and Kim, Sunghwan and Prakash, Alok and Rus, Daniela and Zhao, Jinhua and Low, Bryan Kian Hsiang and Liang, Paul Pu},
	month = jul,
	year = {2025},
	note = {arXiv:2506.15841 [cs]
TLDR: MEM1, an end-to-end reinforcement learning framework that enables agents to operate with constant memory across long multi-turn tasks, is introduced, an end-to-end reinforcement learning framework that enables agents to operate with constant memory across long multi-turn tasks.},
}

@misc{tan_prospect_2025,
	title = {In {Prospect} and {Retrospect}: {Reflective} {Memory} {Management} for {Long}-term {Personalized} {Dialogue} {Agents}},
	shorttitle = {In {Prospect} and {Retrospect}},
	url = {http://arxiv.org/abs/2503.08026},
	doi = {10.48550/arXiv.2503.08026},
	urldate = {2025-08-16},
	publisher = {arXiv},
	author = {Tan, Zhen and Yan, Jun and Hsu, I.-Hung and Han, Rujun and Wang, Zifeng and Le, Long T. and Song, Yiwen and Chen, Yanfei and Palangi, Hamid and Lee, George and Iyer, Anand and Chen, Tianlong and Liu, Huan and Lee, Chen-Yu and Pfister, Tomas},
	month = jul,
	year = {2025},
	note = {arXiv:2503.08026 [cs]
TLDR: Reflective Memory Management (RMM) is proposed, a novel mechanism for long-term dialogue agents, integrating forward- and backward-looking reflections: Prospective Reflection, which dynamically summarizes interactions across granularities-utterances, turns, and sessions-into a personalized memory bank for effective future retrieval, and Retrospective Reflection, which iteratively refines the retrieval in an online reinforcement learning (RL) manner based on LLMs'cited evidence.},
}

@misc{zhong_memorybank_2023,
	title = {{MemoryBank}: {Enhancing} {Large} {Language} {Models} with {Long}-{Term} {Memory}},
	shorttitle = {{MemoryBank}},
	url = {http://arxiv.org/abs/2305.10250},
	doi = {10.48550/arXiv.2305.10250},
	urldate = {2025-06-11},
	publisher = {arXiv},
	author = {Zhong, Wanjun and Guo, Lianghong and Gao, Qiqi and Ye, He and Wang, Yanlin},
	month = may,
	year = {2023},
	note = {arXiv:2305.10250 [cs]
TLDR: MemoryBank incorporates a memory updating mechanism, inspired by the Ebbinghaus Forgetting Curve theory, that permits the AI to forget and reinforce memory based on time elapsed and the relative significance of the memory, thereby offering a more human-like memory mechanism and enriched user experience.},
}

@article{kang_memory_nodate,
	title = {Memory {OS} of {AI} {Agent}},
	language = {en},
	author = {Kang, Jiazheng and Ji, Mingming and Zhao, Zhe and Bai, Ting},
}

@misc{packer_memgpt_2024,
	title = {{MemGPT}: {Towards} {LLMs} as {Operating} {Systems}},
	shorttitle = {{MemGPT}},
	url = {http://arxiv.org/abs/2310.08560},
	doi = {10.48550/arXiv.2310.08560},
	urldate = {2025-06-03},
	publisher = {arXiv},
	author = {Packer, Charles and Wooders, Sarah and Lin, Kevin and Fang, Vivian and Patil, Shishir G. and Stoica, Ion and Gonzalez, Joseph E.},
	month = feb,
	year = {2024},
	note = {arXiv:2310.08560 [cs]
TLDR: This work introduces MemGPT (Memory-GPT), a system that intelligently manages different memory tiers in order to effectively provide extended context within the LLM's limited context window, and utilizes interrupts to manage control flow between itself and the user.},
}

@misc{maharana_evaluating_2024,
	title = {Evaluating {Very} {Long}-{Term} {Conversational} {Memory} of {LLM} {Agents}},
	url = {http://arxiv.org/abs/2402.17753},
	doi = {10.48550/arXiv.2402.17753},
	urldate = {2025-06-03},
	publisher = {arXiv},
	author = {Maharana, Adyasha and Lee, Dong-Ho and Tulyakov, Sergey and Bansal, Mohit and Barbieri, Francesco and Fang, Yuwei},
	month = feb,
	year = {2024},
	note = {arXiv:2402.17753 [cs]
TLDR: A machine-human pipeline is introduced to generate high-quality, very long-term dialogues by leveraging LLM-based agent architectures and grounding their dialogues on personas and temporal event graphs, and presents a comprehensive evaluation benchmark to measure long-term memory in models.},
}

@misc{wu_longmemeval_2025,
	title = {{LongMemEval}: {Benchmarking} {Chat} {Assistants} on {Long}-{Term} {Interactive} {Memory}},
	shorttitle = {{LongMemEval}},
	url = {http://arxiv.org/abs/2410.10813},
	doi = {10.48550/arXiv.2410.10813},
	urldate = {2025-06-03},
	publisher = {arXiv},
	author = {Wu, Di and Wang, Hongwei and Yu, Wenhao and Zhang, Yuwei and Chang, Kai-Wei and Yu, Dong},
	month = mar,
	year = {2025},
	note = {arXiv:2410.10813 [cs]
TLDR: This study introduces LongMemEval, a comprehensive benchmark designed to evaluate five core long-term memory abilities of chat assistants: information extraction, multi-session reasoning, temporal reasoning, knowledge updates, and abstention, and proposes several memory design optimizations including session decomposition for value granularity, fact-augmented key expansion for indexing, and time-aware query expansion for refining the search scope.},
}

@misc{rasmussen_zep_2025,
	title = {Zep: {A} {Temporal} {Knowledge} {Graph} {Architecture} for {Agent} {Memory}},
	shorttitle = {Zep},
	url = {http://arxiv.org/abs/2501.13956},
	doi = {10.48550/arXiv.2501.13956},
	urldate = {2025-05-26},
	publisher = {arXiv},
	author = {Rasmussen, Preston and Paliychuk, Pavlo and Beauvais, Travis and Ryan, Jack and Chalef, Daniel},
	month = jan,
	year = {2025},
	note = {arXiv:2501.13956 [cs]
TLDR: Zep is introduced, a novel memory layer service for AI agents that outperforms the current state-of-the-art system, MemGPT, in the Deep Memory Retrieval (DMR) benchmark and is validated through the more challenging LongMemEval benchmark, which better reflects enterprise use cases through complex temporal reasoning tasks.},
}

@misc{chhikara_mem0_2025,
	title = {Mem0: {Building} {Production}-{Ready} {AI} {Agents} with {Scalable} {Long}-{Term} {Memory}},
	shorttitle = {Mem0},
	url = {https://arxiv.org/abs/2504.19413v1},
	language = {en},
	urldate = {2025-05-23},
	journal = {arXiv.org},
	author = {Chhikara, Prateek and Khant, Dev and Aryan, Saket and Singh, Taranjeet and Yadav, Deshraj},
	month = apr,
	year = {2025},
}

@misc{rasmussen_zep_2025-1,
	title = {Zep: {A} {Temporal} {Knowledge} {Graph} {Architecture} for {Agent} {Memory}},
	shorttitle = {Zep},
	url = {http://arxiv.org/abs/2501.13956},
	doi = {10.48550/arXiv.2501.13956},
	urldate = {2025-05-06},
	publisher = {arXiv},
	author = {Rasmussen, Preston and Paliychuk, Pavlo and Beauvais, Travis and Ryan, Jack and Chalef, Daniel},
	month = jan,
	year = {2025},
	note = {arXiv:2501.13956 [cs]
TLDR: Zep is introduced, a novel memory layer service for AI agents that outperforms the current state-of-the-art system, MemGPT, in the Deep Memory Retrieval (DMR) benchmark and is validated through the more challenging LongMemEval benchmark, which better reflects enterprise use cases through complex temporal reasoning tasks.},
}

@misc{park_generative_2023,
	title = {Generative {Agents}: {Interactive} {Simulacra} of {Human} {Behavior}},
	shorttitle = {Generative {Agents}},
	url = {http://arxiv.org/abs/2304.03442},
	urldate = {2023-08-31},
	publisher = {arXiv},
	author = {Park, Joon Sung and O'Brien, Joseph C. and Cai, Carrie J. and Morris, Meredith Ringel and Liang, Percy and Bernstein, Michael S.},
	month = aug,
	year = {2023},
	note = {arXiv:2304.03442 [cs]},
}

\appendix

\section{Appendix}
\label{sec:appendix}

\begin{table*}[t]
\centering
\setlength{\tabcolsep}{6pt}
\caption{Performance comparison of different embedding models with GPT-4o-mini backbone on \textsc{LoCoMo} dataset.}
\label{tab:embedding_comparison}
\begin{tabular}{lcccccc}
\toprule
\textbf{Embedding Model} & \textbf{ACC} & \textbf{Temporal} & \textbf{Multi-Hop} & \textbf{Open-Domain} & \textbf{Single-Hop}  \\
\midrule
\texttt{text-embedding-v4} & 73.01 & 68.51 & 60.80 & \textbf{59.55} & 80.37 \\
\texttt{text-embedding3-small} & \textbf{76.69} &\textbf{ 71.03} & \textbf{66.67} & 58.33 & \textbf{84.30}  \\
\bottomrule
\end{tabular}
\end{table*}

\begin{figure*}
    \centering
    \includegraphics[width=\textwidth]{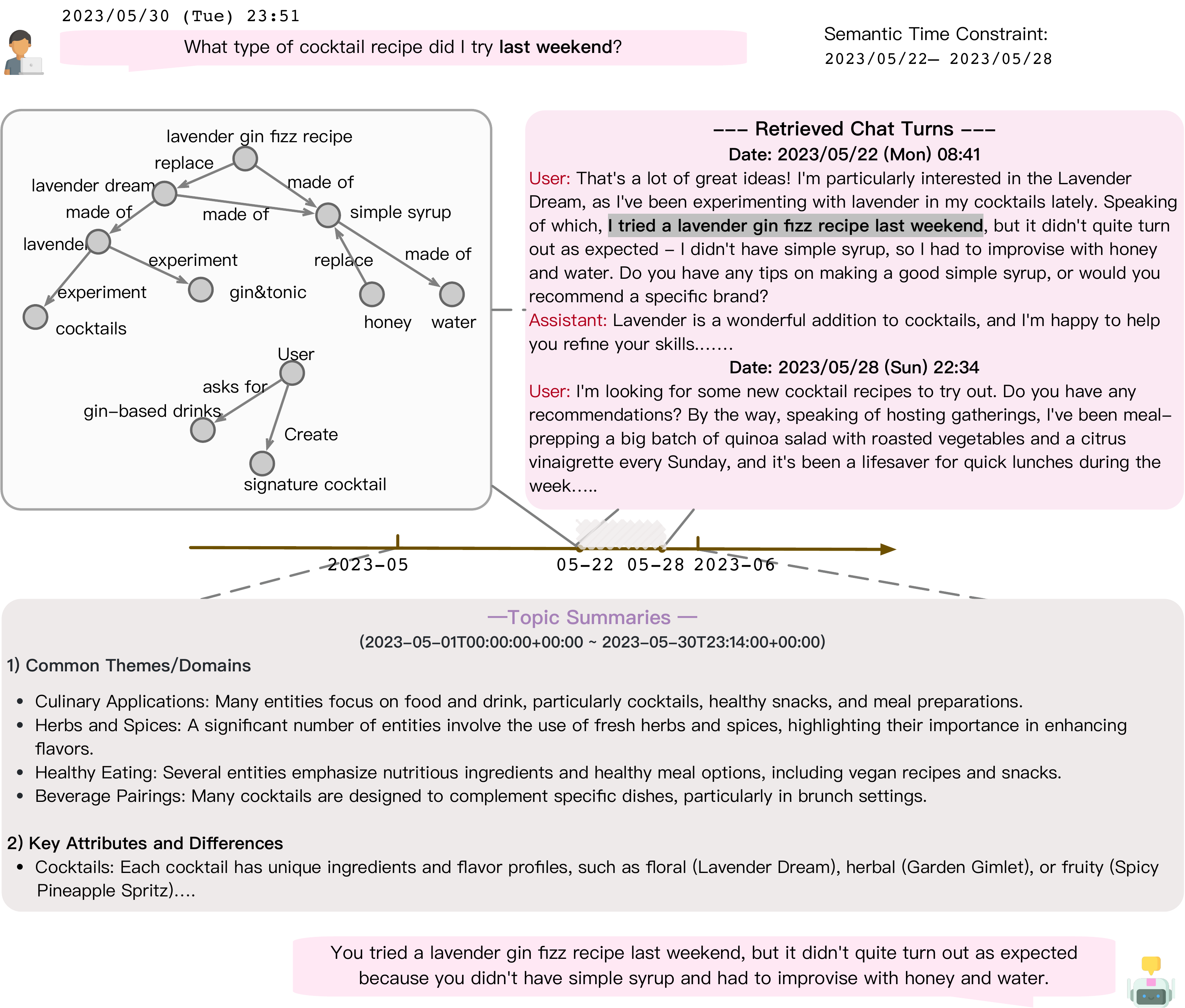} 
    \caption{Case Study of \ours{}.}
    \label{fig:case}
\end{figure*}

\subsection{Implementation Details}
\label{appendix:experimental details}

\subsubsection{Parameter Setup}
We use the following hyper-parameters for all experiments:
\begin{itemize}
    \item \textbf{LLM:} GPT-4o-mini and Qwen3-30B-A3B-Instruct-2507 are used for all stages. During the memory construction stage, we only use the user message for efficiency. During the utilization stage, the generation parameters are:
    \begin{itemize}
        \item Temperature: 0.0
        \item Max tokens: 8192
    \end{itemize}
    \item \textbf{Retriever:} For the embedding model, we use \texttt{text-embedding-3-small} from OpenAI. Additionally, we performed experiments with \texttt{text-embedding-v4} from Qwen. Experimental results are listed in Table~\ref{tab:embedding_comparison}. We use the following configuration: 
    \begin{itemize}
        \item Top-K: 25
    \end{itemize}
    
\end{itemize}

\subsubsection{Hardware}
Experiments are conducted on a machine equipped with 8 Nvidia A100 GPUs, each with 80 GB of RAM. The total available system memory is 256 GB.

\subsection{Case Study}

Figure~\ref{fig:case} illustrates a concrete memory Utilization of \ours{}. In this example, the user asks about a cocktail recipe. Given the query time (2023-05-30) and the temporal expression “last weekend,” SpaCy parses the corresponding semantic time range as 2023-05-22 to 2023-05-28.
Based on this temporal constraint, \ours{} first queries the TKG to retrieve all facts whose valid time falls within the identified interval.
Then, \ours{} performs dense retrieval over both raw dialogue and summaries. Only summaries whose temporal scope satisfies the query’s time constraint are considered. The retrieved facts from the TKG are then used as contextual signals to rerank the candidate chunks, promoting those that are semantically aligned with the time constraint. As a result, the chunk containing the correct cocktail recipe is ranked at the top.

During the graph retrieval, although several related facts are found, none of them can directly answer the user’s question. This highlights a key limitation of using TKG facts alone as ground-truth memory: while they capture structured and time-aware information, they contain insufficient, point-wise, and instant knowledge.

Overall, \ours{} effectively combines temporal reasoning and semantic retrieval to produce accurate and temporally aligned responses, demonstrating its advantage over methods that rely solely on timestamped facts or unfiltered dense retrieval.

\subsection{Datasets and Baseliens}
\paragraph{Datasets.}
The \textsc{LongMemEval} dataset is a comprehensive, challenging, and scalable benchmark for testing the long-term memory of chat assistants.
Two standard test sets are created for 500 questions:
\textsc{LongMemEval\_S} with each question's chat history has roughly 115k tokens (30-40 sessions) and \textsc{LongMemEval\_M}: each question's chat history has roughly 500 sessions (~1.5M tokens).
In our work, we adopt the \textsc{LongMemEval-S} version due to its balance between dialogue length and computational feasibility.

The \textsc{LoCoMo} benchmark targets the evaluation of long-range conversational memory. It features extremely long dialogues, with each conversation spanning roughly 300 turns and around 9K tokens on average. Note that questions in the \textsc{LoCoMo} dataset do not contain explicit query timestamps. We use the session start time as the reference timestamp when extracting temporal constraints from queries.

\paragraph{Baselines.}
 (1) LangMem~\cite{langchain_langchain_nodate} is the Langchain's long-term memory module.\\
 (2) A-MEM~\cite{xu2025mem} system dynamically structures memories through notes. Each note has attributes like keywords, contextual descriptions, and tags generated by the LLM. Retrieval from memory is conducted through semantic similarity.\\
 (3) MemoryOS~\citep{kang_memory_nodate} organizes conversational memory in an OS-inspired hierarchy, structuring interactions into short-term, mid-term, and long-term layers via paging and heat-based updating.\\
 (4) Mem0 ~\citep{chhikara2025mem0} extracts memories from dialogue turns through a combination of global summaries and recent context, maintaining them via LLM-guided operations. Mem0$^g$ further propose an enhanced variant that leverages graph-based memory representations to capture complex relational structures among conversational elements.\\
 (5) Zep~\citep{rasmussen_zep_2025} is a temporal knowledge graph architecture that organizes data into episodic, semantic, and community subgraphs to capture dynamic, time-sensitive relationships. 
For comparison, we include the baseline results from their original papers and LightMem~\cite{fang_lightmem_2025}. Due to computational resource constraints, all experiments are conducted with a single run using a fixed setting. 
\subsection{LLM-as-Judge prompts}
\label{prompts:LLM-as-Judge}

\begin{table*}[!b]  
\begin{tcolorbox}[
    colback=black!5, 
    colframe=black!70!white, 
    title=Standard Tasks (Single-session-user/assistant\, Multi-session) for LongMemEval Dataset,
    fonttitle=\bfseries,
    arc=3mm, 
    boxrule=0.8pt 
]
\begin{tabularx}{\textwidth}{X}
I will give you a question, a correct answer, and a response from a model. Please answer yes if the response contains the correct answer. Otherwise, answer no. If the response is equivalent to the correct answer or contains all the intermediate steps to get the correct answer, you should also answer yes. If the response only contains a subset of the information required by the answer, answer no.

\textbf{Question:} \{question\}\\
\textbf{Correct Answer:} \{answer\}\\
\textbf{Model Response:} \{response\}

Is the model response correct? Answer yes or no only.
\end{tabularx}
\end{tcolorbox}
\end{table*}

\begin{table*}[!b]  
\begin{tcolorbox}[
    colback=black!5, 
    colframe=black!70!white, 
    title=Temporal Reasoning Tasks for LongMemEval Dataset,
    fonttitle=\bfseries,
    arc=3mm, 
    boxrule=0.8pt 
]
\begin{tabularx}{\textwidth}{X}
I will give you a question, a correct answer, and a response from a model. Please answer yes if the response contains the correct answer. Otherwise, answer no. If the response is equivalent to the correct answer or contains all the intermediate steps to get the correct answer, you should also answer yes. If the response only contains a subset of the information required by the answer, answer no. In addition, do not penalize off-by-one errors for the number of days. If the question asks for the number of days/weeks/months, etc., and the model makes off-by-one errors (e.g., predicting 19 days when the answer is 18), the model's response is still correct.

\textbf{Question:} \{question\}\\
\textbf{Correct Answer:} \{answer\}\\
\textbf{Model Response:} \{response\}

Is the model response correct? Answer yes or no only.
\end{tabularx}
\end{tcolorbox}
\end{table*}

\begin{table*}[!b]  
\begin{tcolorbox}[
    colback=black!5, 
    colframe=black!70!white, 
    title=Knowledge Update Tasks for LongMemEval Dataset,
    fonttitle=\bfseries,
    arc=3mm, 
    boxrule=0.8pt 
]
\begin{tabularx}{\textwidth}{X}
I will give you a question, a correct answer, and a response from a model. Please answer yes if the response contains the correct answer. Otherwise, answer no. If the response contains some previous information along with an updated answer, the response should be considered as correct as long as the updated answer is the required answer.

\textbf{Question:} \{question\}\\
\textbf{Correct Answer:} \{answer\}\\
\textbf{Model Response:} \{response\}

Is the model response correct? Answer yes or no only.
\end{tabularx}
\end{tcolorbox}
\end{table*}

\begin{table*}[!b]  
\begin{tcolorbox}[
    colback=black!5, 
    colframe=black!70!white, 
    title=Single-session Preference Tasks for LongMemEval Dataset,
    fonttitle=\bfseries,
    arc=3mm, 
    boxrule=0.8pt 
]
\begin{tabularx}{\textwidth}{X}
I will give you a question, a rubric for desired personalized response, and a response from a model. Please answer yes if the response satisfies the desired response. Otherwise, answer no. The model does not need to reflect all the points in the rubric. The response is correct as long as it recalls and utilizes the user's personal information correctly.

\textbf{Question:} \{question\}\\
\textbf{Rubric:} \{answer\}\\
\textbf{Model Response:} \{response\}

Is the model response correct? Answer yes or no only.
\end{tabularx}
\end{tcolorbox}
\end{table*}

\begin{table*}[!b]  
\begin{tcolorbox}[
    colback=black!5, 
    colframe=black!70!white, 
    title=Abstention Tasks for LongMemEval Dataset,
    fonttitle=\bfseries,
    arc=3mm, 
    boxrule=0.8pt 
]
\begin{tabularx}{\textwidth}{X}
I will give you an unanswerable question, an explanation, and a response from a model. Please answer yes if the model correctly identifies the question as unanswerable. The model could say that the information is incomplete, or some other information is given but the asked information is not.

\textbf{Question:} \{question\}\\
\textbf{Explanation:} \{answer\}\\
\textbf{Model Response:} \{response\}

Does the model correctly identify the question as unanswerable? Answer yes or no only.
\end{tabularx}
\end{tcolorbox}
\end{table*}

\begin{table*}[!b]  
\begin{tcolorbox}[
    colback=black!5, 
    colframe=black!70!white, 
    title=LoCoMo Dataset,
    fonttitle=\bfseries,
    arc=3mm, 
    boxrule=0.8pt 
]
\begin{tabularx}{\textwidth}{X}
Your task is to label an answer to a question as 'CORRECT' or 'WRONG'. You will be given the following data:\\
\hspace{2em}(1) a question (posed by one user to another user),\\
\hspace{2em}(2) a 'gold' (ground truth) answer,\\
\hspace{2em}(3) a generated answer\\
which you will score as CORRECT/WRONG.\\\\

The point of the question is to ask about something one user should know about the other user based on their prior conversations.\\
The gold answer will usually be a concise and short answer that includes the referenced topic, for example:\\
Question: Do you remember what I got the last time I went to Hawaii?\\
Gold answer: A shell necklace\\
The generated answer might be much longer, but you should be generous with your grading - as long as it touches on the same topic as the gold answer, it should be counted as CORRECT.\\
\\
For time related questions, the gold answer will be a specific date, month, year, etc. The generated answer might be much longer or use relative time references (like "last Tuesday" or "next month"), but you should be generous with your grading - as long as it refers to the same date or time period as the gold answer, it should be counted as CORRECT. Even if the format differs (e.g., "May 7th" vs "7 May"), consider it CORRECT if it's the same date.\\

Now it's time for the real question:\\\\

\textbf{Question:} \{question\}\\
\textbf{Gold answer:} \{gold\_answer\}\\
\textbf{Generated answer:} \{generated\_answer\}\\\\

First, provide a short (one sentence) explanation of your reasoning, then finish with CORRECT or WRONG.
Do NOT include both CORRECT and WRONG in your response, or it will break the evaluation script.\\\\

Just return the label CORRECT or WRONG in a json format with the key as "label".
\end{tabularx}
\end{tcolorbox}
\end{table*}

\end{document}